\documentclass[sigconf]{acmart}

\AtBeginDocument{%
  \providecommand\BibTeX{{%
    \normalfont B\kern-0.5em{\scshape i\kern-0.25em b}\kern-0.8em\TeX}}}

\copyrightyear{2022}
\acmYear{2022}
\setcopyright{acmlicensed}\acmConference[WWW '22]{Proceedings of the ACM
Web Conference 2022}{April 25--29, 2022}{Virtual Event, Lyon, France}
\acmBooktitle{Proceedings of the ACM Web Conference 2022 (WWW '22), April
25--29, 2022, Virtual Event, Lyon, France}
\acmPrice{15.00}
\acmDOI{10.1145/3485447.3512070}
\acmISBN{978-1-4503-9096-5/22/04}

\usepackage{array,enumitem,multirow,caption,graphicx,subcaption,multicol,lipsum,float}
\usepackage{amsmath, amsfonts, amsthm}
\usepackage[ruled,vlined,linesnumbered]{algorithm2e}
\usepackage{kotex} 
\usepackage{makecell}
\usepackage{color, colortbl}
\usepackage{color}
\usepackage{array,enumitem,multirow,caption,graphicx,subcaption,multicol,lipsum,float,adjustbox}
\usepackage{amsmath, amsfonts, amsthm}
\usepackage[colorinlistoftodos,textsize=tiny]{todonotes}
\newcolumntype{C}[1]{>{\centering\let\newline\\\arraybackslash\hspace{0pt}}m{#1}}
\newcolumntype{Z}{>{\centering\arraybackslash}m{0.062\linewidth}}
\usepackage[compatible]{algpseudocode} 
\renewcommand{\algorithmiccomment}[1]{\bgroup\hfill\small//~#1\egroup}

\begin{document}

\title{Consensus Learning from Heterogeneous Objectives \\for One-Class Collaborative Filtering}
\fancyhead{}

\author{SeongKu Kang$^1$, Dongha Lee$^2$, Wonbin Kweon$^1$, Junyoung Hwang$^1$, Hwanjo Yu$^1$\*}
\affiliation{%
   \institution{$^1$Pohang University of Science and Technology (POSTECH), South Korea}
   \institution{$^2$University of Illinois at Urbana-Champaign (UIUC), United States}
   \country{}
   \{seongku, kwb4453, jyhwang, hwanjoyu\}@postech.ac.kr, donghal@illinois.edu
}
\authornotemark[0]
\authornote{corresponding author}
\renewcommand{\shortauthors}{SeongKu Kang et al.}

\begin{abstract}
Over the past decades, for One-Class Collaborative Filtering (OCCF), many learning objectives have been researched based on a variety of underlying probabilistic models.
From our analysis, we observe that models trained with different OCCF objectives capture distinct aspects of user-item relationships, which in turn produces complementary recommendations.
This paper proposes a novel OCCF framework, named as \proposed, that exploits the complementarity from heterogeneous objectives throughout the training process, generating a more generalizable model.
\proposed constructs a multi-branch variant of a given target model by adding auxiliary heads, each of which is trained with heterogeneous objectives.
Then, it generates \textit{consensus} by consolidating the various views from the heads, and guides the heads based on the consensus.
The heads are collaboratively evolved based on their complementarity throughout the training, which again results in generating more accurate consensus iteratively.
After training, we convert the multi-branch architecture back to the original target model by removing the auxiliary heads, thus there is no extra inference cost for the deployment.
Our extensive experiments on real-world datasets demonstrate that \proposed significantly improves the generalization of the model by exploiting the complementarity from heterogeneous objectives.
\end{abstract}


\newcommand{\proposed}{ConCF\xspace}
\newcommand{\bl}{SingleCF\xspace}
\maketitle

\section{Introduction}

One-Class Collaborative Filtering (OCCF) aims to discover users' preferences and recommend the items that they might like in the future, given a set of observed user-item interactions (e.g., clicks or purchases) \cite{pan2008one, BUIR}.
To effectively learn users' preferences from such implicit feedback, many learning objectives (or objective functions) have been researched based on a variety of underlying probabilistic models.
In particular, pair-wise ranking objective \cite{BPR}, metric-learning objective \cite{CML}, and binary cross-entropy \cite{NeuMF} have shown good performance.
However, there is no absolute winner among them that can always achieve the best performance, because their superiority varies depending on datasets, model architectures, and evaluation metrics \cite{sun2020we}.
Since empirical comparison of all the choices is exceedingly costly in terms of both computing power and time consumption, most existing work simply leverages a \textit{generally good} one for their model optimization.


In this paper, we analyze OCCF models optimized by various learning objectives and observe that \emph{models trained with different objectives capture distinct aspects of user-item relationships.}
We observe that the test interactions (i.e., the ground truth of the users' preference) correctly predicted\footnote{We consider the test interactions included in the top-$N$ ranking list as correct predictions, also known as top-$N$ recommendation \cite{BUIR}.}
by each model are significantly different, regardless of their quantitative recommendation performance;
a model with low performance still correctly predicts considerable portions of the test interactions that the other models predict incorrectly.
Further, we demonstrate that the different and complementary knowledge induced by heterogeneous objectives can be exploited to provide more accurate recommendations to a larger number of users, compared to the case of considering each single-faceted knowledge.
The observations lead us to exploit the complementary knowledge for training a model to have a more complete understanding of user-item relationships.

\begin{figure}[t]
\begin{subfigure}[t]{0.99\linewidth}
    \includegraphics[width=\linewidth]{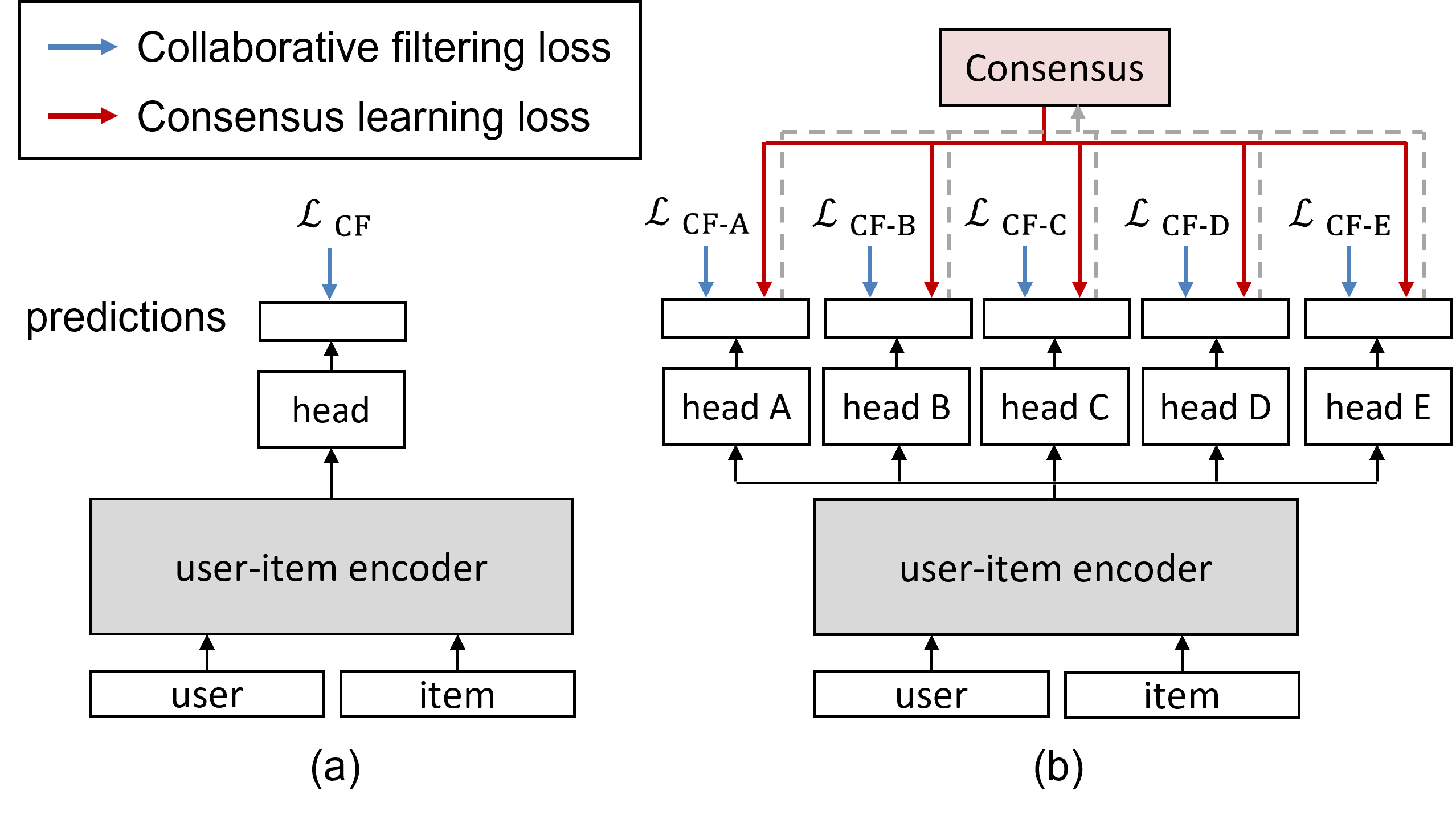}
\end{subfigure}
\vspace{-0.2cm}
\caption{The conceptual illustrations of (a) the existing learning scheme and (b) the proposed framework. 
Unlike (a) that uses a given objective function, (b) exploits multi-faceted knowledge induced by the heterogeneous learning objectives, enabling to train a more generalizable model.}
\label{fig:overview}
\vspace{-0.65cm}
\end{figure}

We propose a new end-to-end OCCF framework, named as \proposed (\underline{Con}sensus learning for OC\underline{CF}), that exploits the complementarity from heterogeneous objectives throughout the training process, generating a more generalizable model (Figure \ref{fig:overview}).
\proposed constructs a multi-branch variant of a given target model by adding auxiliary heads, each of which is trained with different objective functions. 
Then, it generates \textit{consensus} of multiple views from different heads, and guides the heads based on the consensus.
Concretely, each head is trained with two loss terms -- (1) the original collaborative filtering loss, and (2) a consensus learning loss that matches its prediction to the consensus.
The consensus, which consolidates the predictions from differently optimized heads, contains rich and accurate information not sufficiently discovered by a single head.
With the guidance, the heads are collaboratively evolved based on their complementarity during training, which again results in generating more accurate consensus iteratively.

The consensus learning, exploiting the complementarity from heterogeneous learning objectives, poses three unique challenges:

\vspace{0.05cm} \noindent
\textbf{C1}:
Since each objective is based on different probabilistic models, the distributions and semantics of their prediction scores are different.
Thus, the predictions from different heads need to be represented as a unified form so that they can learn interchangeably from each other.
For example, BPR objective \cite{BPR} encodes user-item relations as unbounded scores where larger values indicate higher relevance, whereas CML objective \cite{CML} learns the unit-ball metric space where user-item relations are encoded as their Euclidean distances in the range of [$0, 2$].
So, a score `0' for a user-item pair implies a weak relation in the former, but a very strong relation in the latter.
These discrepancies make the naive score-based learning \cite{DML, BD, ONE} among the heads inapplicable.

\vspace{0.1cm} \noindent
\textbf{C2}:
To generate informative consensus beneficial for all heads, it is essential to identify \textit{reliable} predictions from each head and selectively reflect them into the consensus.
As shown in our analysis, the user-item relationships correctly captured by each objective are significantly different.
Thus, only the reliable predictions correctly captured by each head should be included in the consensus.
Without considering the reliability, the consensus will contain incorrect predictions, providing inaccurate supervision for the heads.

\vspace{0.1cm} \noindent    
\textbf{C3}:
To exploit the complementarity, heterogeneous OCCF objectives having different convergence behaviors need to be properly balanced so that model parameters converge to useful features across all heads.
In case that some of the heads are too dominant during training, it incurs the imbalance that impedes the remaining heads from being fully optimized. 
This results in the degradation of the quality of the consensus and the overall effectiveness~of~\proposed.

\vspace{0.1cm}
\proposed introduces solutions to the aforementioned challenges:
\textbf{First}, to cope with the discrepancy among the outputs of the heads, \proposed utilizes \textit{ranking information}, a universal form of knowledge that can be derived from all heads.
Concretely, \proposed uses the information of relative preference priorities (e.g., user $u$ prefers item $i$ over item $j$.) revealed from the predictions to exchange the knowledge among the heads.
Utilizing the ranking information also has strengths in the top-$N$ recommendation, which provides a short item ranking list \cite{DERRD}.
\textbf{Second}, to generate informative consensus, \proposed takes into account not only the predicted rankings but also their reliability revealed from \textit{temporal consistency} of the predictions.
The temporal consistency measures how consistent the predicted ranking is during training, and according to our findings, this is an important indicator to identify reliable predictions.
By selectively reflecting the temporal-consistent predictions from each head, \proposed generates accurate consensus that every head can agree on, providing informative supervision for all heads.
\textbf{Third}, for balancing the heads optimized in a different way, we enforce all the heads to be trained at similar rates by dynamically adjusting the gradient scale of each head on the shared parameters.
We apply the gradient normalization technique \cite{gradnorm}, enabling the shared parameters to converge to beneficial features across all heads without additional hyperparameters for controlling~the~effects~of~each~heads.

In test time, we convert the multi-branch architecture back to the original single-branch model by removing the auxiliary heads, thus there is no additional inference cost for deployment.
Further, the consensus from multiple heads can be also used as a high capacity model in the scenario where there is less constraint on the inference cost.
The key contributions of our work are as follows:
\vspace{-0.1cm}
\begin{itemize}[leftmargin=*]
    \item Through our extensive analyses, we address the necessity of consolidating multiple views from heterogeneous OCCF objectives, which has not been studied well in the previous literature.
    
    \item We propose \proposed that exploits the complementarity from heterogeneous objectives throughout the training process.
    \proposed effectively deals with the challenges of the consensus learning from the heterogeneous objectives having distinct nature.
    
    \item We validate the superiority of \proposed by extensive experiments on real-world datasets. 
    \proposed achieves superior performance compared to the model optimized by a single objective.
    Also, \proposed generates a more generalizable model in a single-stage process than the alternative strategies of the conventional two-stage knowledge distillation or ensemble models.    
    
\end{itemize}

\begin{figure*}[t]
\centering
\hspace{-0.7cm}
    \begin{minipage}[r]{0.75\linewidth}
        \vspace{0.2cm}
        \centering
        \begin{subfigure}{0.28\textwidth}
            \includegraphics[width=1\textwidth]{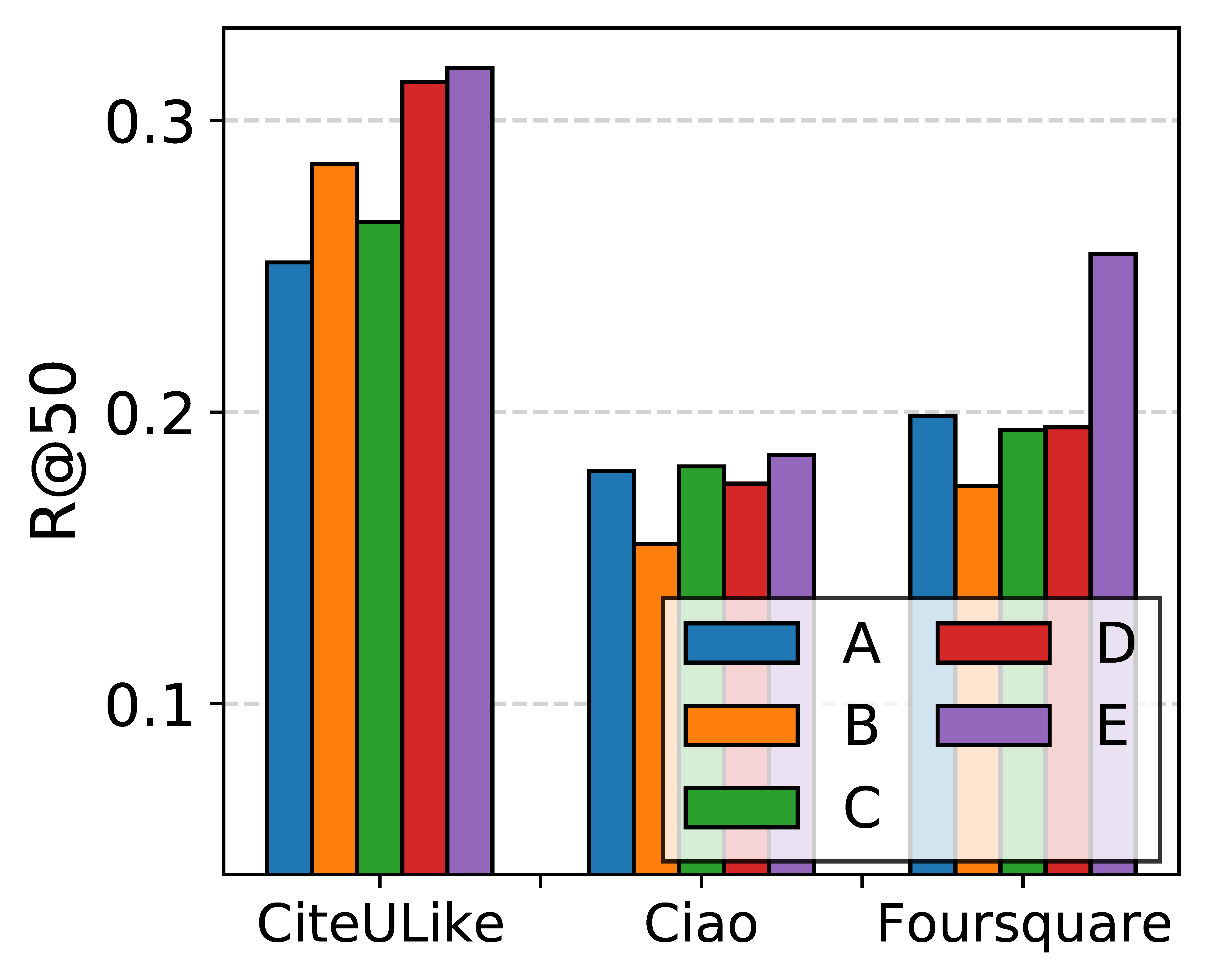}
        \end{subfigure}
        \hspace{-0.23cm}
        \begin{subfigure}{0.23\textwidth}
            \includegraphics[width=1\textwidth]{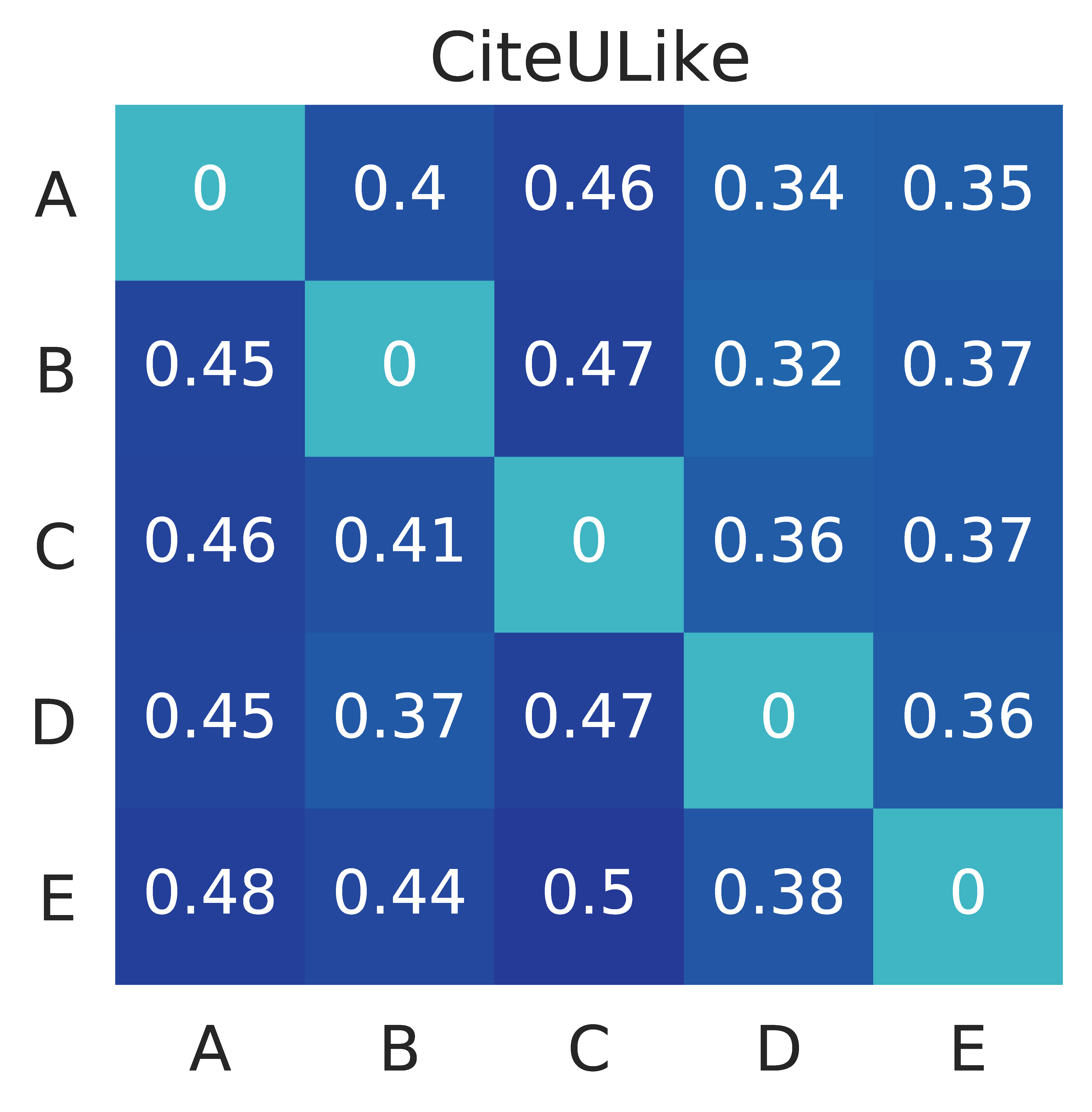}
        \end{subfigure}
        \hspace{-0.25cm}
        \begin{subfigure}{0.23\textwidth}
         \includegraphics[width=1\textwidth]{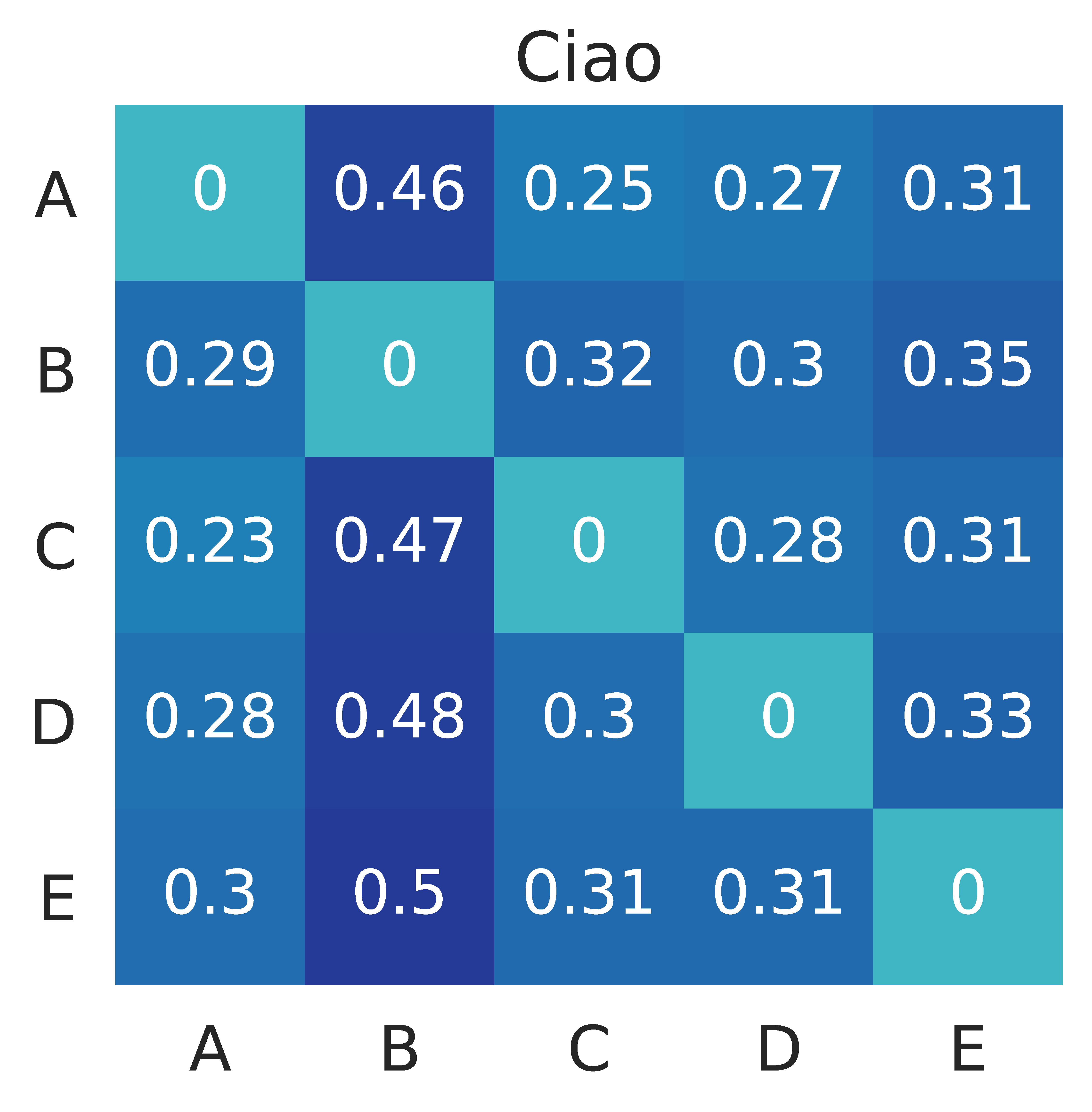}
        \end{subfigure}
        \hspace{-0.25cm}
        \begin{subfigure}{0.28\textwidth}
          \includegraphics[width=1\textwidth]{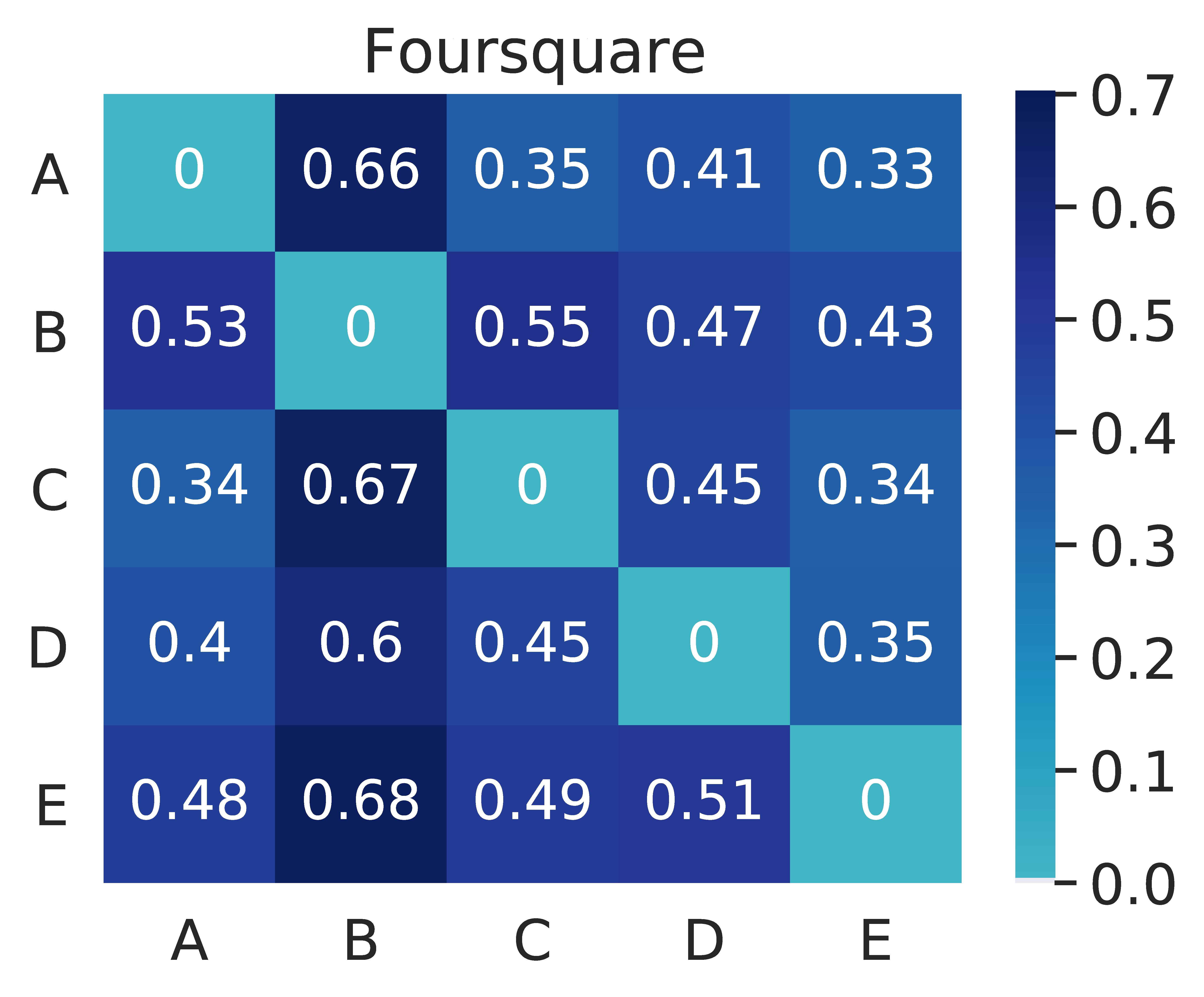}
        \end{subfigure}
        \vspace{-0.1cm}
        \subcaption{Recommendation performance (Recall@50) and PER($x$: row  ;  $y$: column) maps.}
    \end{minipage}
    \hspace{-0.2cm}
    \begin{minipage}[r]{0.23\linewidth}
        \centering
        \begin{subfigure}{1.15\textwidth}
          \includegraphics[width=1\textwidth]{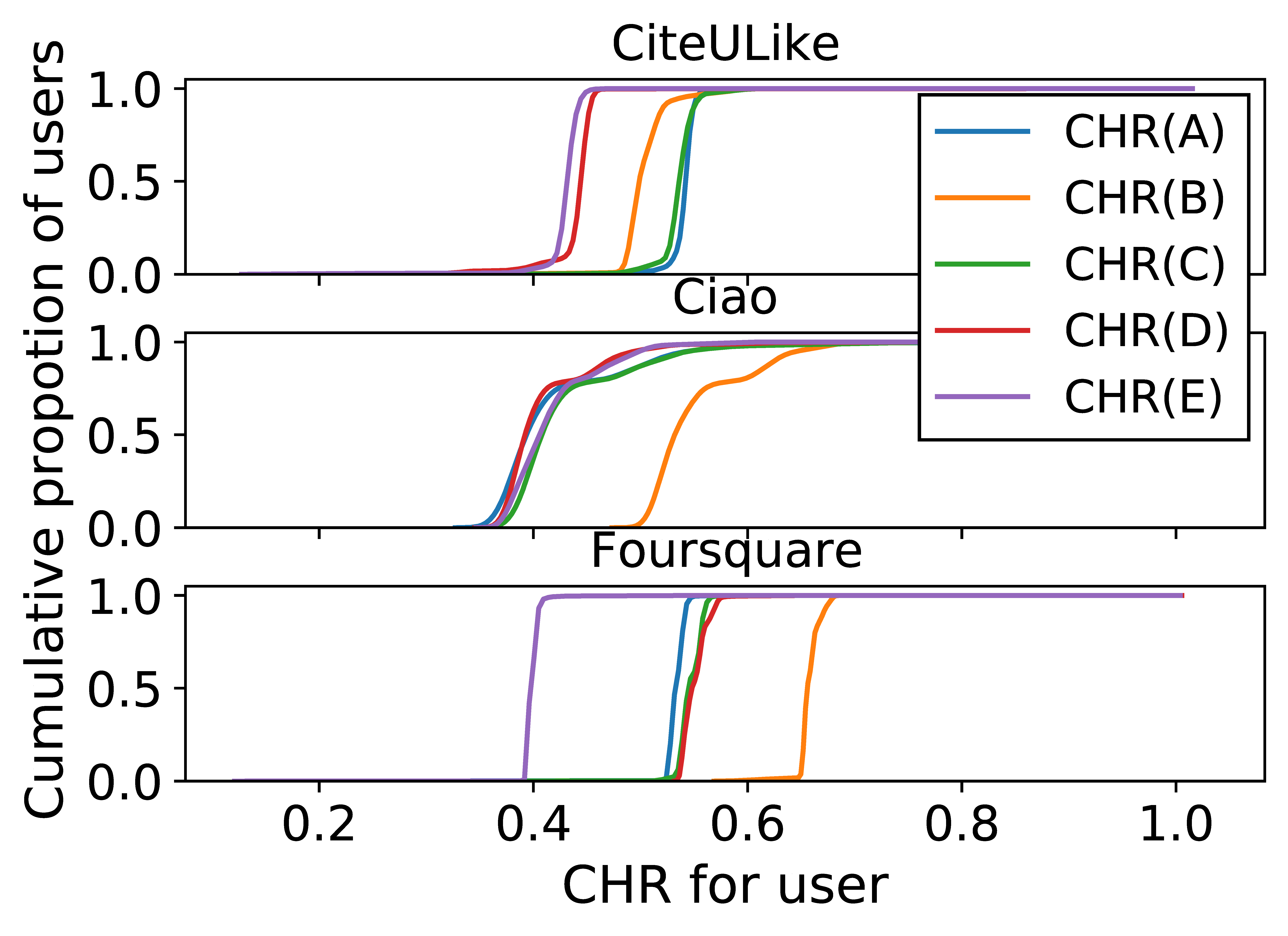}
        \end{subfigure}
        \vspace{-0.2cm}
        \subcaption{The CDF of CHR($y$; $\mathcal{M}_{O}$)}
    \end{minipage}
    \caption{Analyses of complementarity from different learning objectives. A model $x$ is trained with learning objective CF-$x$.}
    \label{fig:prelim_analysis}
    \vspace{-0.4cm}
\end{figure*}

\vspace{-0.1cm}
\section{Preliminaries}
\label{sec:preliminary}

\subsection{Problem Formulation}
Let $\mathcal{U}$ and $\mathcal{I}$ denote a set of users and a set of items, respectively.
Given implicit user–item interactions (e.g., click, purchase), we build a binary matrix $\mathbf{R} \in \{0,1\}^{|\mathcal{U}| \times |\mathcal{I}|}$, where $r_{ui}=1$ if the user $u$ has interacted with the item $i$, otherwise $r_{ui}=0$.
Let $\mathcal{O}^{+}=\left\{(u, i) \mid r_{u i}=1\right\}$ denote a set of observed user-item interactions, and let $\mathcal{O}^{-}=\left\{(u, i) \mid r_{u i}=0\right\}$ denote a set of unobserved user-item pairs.
Also, let $\widetilde{\mathcal{O}}= \{ (u,i,j) \mid r_{u i}=1 \wedge r_{u j}=0 \}$ denote a set of triples $(u,i,j)$ such that user $u$ has interacted with $i$ but has not interacted with $j$.  
The goal of one-class collaborative filtering (OCCF) is to obtain the relevance score $\hat{r}_{ui} \in \mathbb{R}$ indicating how likely the user would interact with (or prefers to) the item $i$.
Based on the predicted scores, we recommend top-$N$ unobserved items with the highest scores for each user, called as~top-$N$~recommendation.

\vspace{-2pt}
\subsection{Learning Objectives for OCCF}
We select five learning objectives for OCCF that have been widely adopted in recent work.
We briefly review their key concepts and underlying probabilistic models.
In case that there are lots of variants, we focus on the representative basic form.
Note that any OCCF objective can be flexibly adopted in the proposed framework.
In this regard, we use alphabets to denote the objectives throughout the paper, i.e., CF-$x$, $x \in \{A, B, C, D, E\}$.

Let $\hat{r}_{ui}$ denote a relevance score of $(u, i)$ pair predicted by a model (Figure \ref{fig:overview}a), and let $\theta$ denote the model parameters.

\noindent
\textbf{CF-A.} 
Bayesian personalized ranking (BPR) \cite{BPR} has been broadly used for model optimization \cite{BPR2021a, BPR2020a, BPR2021b, ADBPR, he2020lightgcn, NGCF}.
It maximizes the posterior probability $p(\theta \mid \widetilde{\mathcal{O}}) \propto p(\widetilde{\mathcal{O}} \mid \theta) p(\theta)$.
The probability that user $u$ prefers item $i$ to item $j$ is defined as $\sigma (\hat{r}_{ui}-\hat{r}_{uj})$, where $\sigma(\cdot)$ is the sigmoid function and $\hat{r}_{ui}$ has unbounded range. 
By taking the negative log-posterior, the loss function is formulated as follows:
\begin{equation}
    \begin{aligned}
        \mathcal{L}_{CF-A} = -\sum_{(u,i,j) \in \widetilde{\mathcal{O}}} \log \sigma (\hat{r}_{ui}-\hat{r}_{uj}). 
    \end{aligned}
\end{equation}
The prior $p(\theta)$ is computed as $\lVert \theta \rVert^2$ by adopting a normal distribution.
We omit the prior from Eq.1 for simplicity.

\vspace{0.02cm} \noindent
\textbf{CF-B.}
Many recent studies have adopted metric learning \cite{CML, feng2015personalized, zhou2019collaborative, tay2018latent, li2020dynamic, li2020symmetric}.
The strength of the user’s preference on the item is considered inversely (or negatively) proportional to their Euclidean distance in the representation space, i.e., $\hat{r}_{ui}= -\lVert \mathbf{u}-\mathbf{i}\rVert_2$.
One of the most representative work \cite{CML} adopts triplet loss as follows:
\begin{equation}
    \begin{aligned}
        \mathcal{L}_{CF-B} = \sum_{(u,i,j) \in \widetilde{\mathcal{O}}} [- \hat{r}_{ui} + \hat{r}_{uj} + m]_+,
    \end{aligned}
\end{equation}
where $[x]_+ = max(x, 0)$ is the hinge loss, and $m$ is the margin.
The space has a unit-ball constraint that enforces the size of all user and item representations less than or equal to 1.

\vspace{0.02cm} \noindent
\textbf{CF-C.}
Binary cross-entropy is a highly popular choice for model optimization \cite{BCE2020a, NeuMF, BCE2020b, BCE2020c, BCE2020d, NCR}.
It assumes the observations are sampled from the Bernoulli distribution, i.e., 
$p(\mathcal{O}^+, \mathcal{O}^- \mid \theta) = \prod_{(u,i) \in \mathcal{O}^+ \cup \mathcal{O}^-} {\hat{r}_{ui}}^{r_{ui}}{(1-\hat{r}_{ui})}^{(1-r_{ui})}$ where $\hat{r}_{ui} \in [0,1]$.
By taking the negative log-likelihood, the loss function is defined as follows:
\begin{equation}
    \begin{aligned}
        \mathcal{L}_{CF-C} = -\sum_{(u,i) \in \mathcal{O}^+ \cup  \mathcal{O}^-} r_{ui} \log \hat{r}_{ui} + (1-r_{ui}) \log(1- \hat{r}_{ui})
    \end{aligned}
\end{equation}

\vspace{0.02cm} \noindent
\textbf{CF-D.} Another popular choice is to utilize Gaussian distribution for the observation sampling distribution \cite{PMF, SQR2020a, hu2008collaborative, CDAE, VAE}, i.e., $p(\mathcal{O}^+, \mathcal{O}^- \mid \theta) = \prod_{(u,i) \in \mathcal{O}^+ \cup \mathcal{O}^-} \mathcal{N}(r_{ui} \mid \hat{r}_{ui}, \sigma^2)$, with a fixed variance $\sigma^2$. The score $\hat{r}_{ui}$ has unbounded range.
By taking the negative log-likelihood, the loss function is defined as follows:
\begin{equation}
    \begin{aligned}
        \mathcal{L}_{CF-D} = \frac{1}{2} \sum_{(u,i) \in \mathcal{O}^+ \cup  \mathcal{O}^-}  \left( r_{ui} -\hat{r}_{ui}\right)^2
    \end{aligned}
\end{equation}


\vspace{0.02cm} \noindent
\textbf{CF-E.} The multinomial distribution, which is widely used in~language models \cite{blei2003latent}, has been adopted in many recent work \cite{VAE, MUL2020a, MUL2020b, MUL2020c}. 
It assumes each user's observations are sampled as
$p(\mathcal{O}^+_u, \mathcal{O}^-_u \mid \theta) = \frac{\Gamma(\sum_{i\in \mathcal{I}} r_{ui}+1)}{\prod_{i\in \mathcal{I}}\Gamma(r_{ui}+1)} \prod_{i\in \mathcal{I}}{\hat{r}_{ui}}^{r_{ui}} $ where $\sum_{i \in \mathcal{I}} \hat{r}_{ui} =1$ and $\hat{r}_{ui} \in [0,1]$.
By taking the negative log-likelihood, the loss function is defined as follows:
\begin{equation}
    \begin{aligned}
        \mathcal{L}_{CF-E} = -\sum_{(u,i) \in \mathcal{O}^+ \cup  \mathcal{O}^-} r_{ui}  \log \hat{r}_{ui}
    \end{aligned}
\end{equation}

\noindent
\textbf{Remarks.}
Most previous studies use a single OCCF objective for model optimization.
However, the heterogeneous objectives, which are based on different assumptions and probabilistic models, inject different inductive biases into the model \cite{inductive2}.
These inductive biases make the model prefer some hypotheses over other hypotheses, and thus better at solving certain problems and worse at others \cite{inductive2}.
In the next section, we investigate the complementarity of knowledge induced by heterogeneous learning~objectives.

\subsection{Analyses of Complementarity induced by Heterogeneous Learning Objectives}
We provide in-depth analyses of the knowledge induced by heterogeneous objectives on three real-world datasets.
We independently train five models (Fig.\ref{fig:overview}a) with each CF function in Section 2.2,
and their performances are summarized in the leftmost chart in Fig.\ref{fig:prelim_analysis}a.
All other controllable factors except for the loss function are fixed.
See Appendix \ref{app:setup} for details of experiment setup.

\subsubsection{Pairwise Exclusive-hit Ratio (PER)}
$\text{PER}(x;y)$ quantifies the knowledge of user-item relationship captured by the model $x$ but failed to be effectively captured by the model $y$.
Let $\mathcal{H}_{x}$ denote the set of test interactions correctly predicted\footnote{We assume that the knowledge captured by a model is revealed by its predictions. Here, we consider the test items (i.e., the ground truth of the user’s preference) that are in the top-50 ranking list as the correct predictions.} by the model $x$.
We define PER as follows:
\begin{equation}
    \begin{aligned}
        \text {PER} (x; y) =\frac{| \mathcal{H}_{x}  - \mathcal{H}_{y}|}{|\mathcal{H}_{x}|}.
    \end{aligned}
\end{equation}
A high PER value means that the two models capture significantly different aspects from the same dataset.
We compute PER for all pairs of models trained with each learning objective, and the PER maps are summarized in Figure \ref{fig:prelim_analysis}a.

\vspace{0.02cm}
\noindent
\textbf{Observations.} The models trained with different objectives have considerably high PER overall.
Even the model with low performance still better predicts considerable portions of the test interactions than the other models with much higher performance, 
e.g., on CiteULike, the model trained by CF-E achieves 26\% higher performance than that by CF-A, but it fails to correctly predict 35\% of the test interactions correctly predicted by the latter, i.e.,~PER(A;E)=0.35.

\subsubsection{Complementary Hit Ratio (CHR)}
We investigate the complementarity that a model can learn from the other models.
Given a set of trained models $\mathcal{M}$, we define CHR for a model $y$ as follows:
\begin{equation}
    \begin{aligned}
        \text {CHR} (y; \mathcal{M}) = \frac{|\bigcup_{x \in \mathcal{M}}\mathcal{H}_x - \mathcal{H}_y|}{|\bigcup_{x \in \mathcal{M}} \mathcal{H}_x|}.
    \end{aligned}
    \label{eq:chr}
\end{equation}
$\bigcup_{x \in \mathcal{M}} \mathcal{H}_x$ is the total test interaction set correctly predicted by~$\mathcal{M}$.
CHR($y; \mathcal{M}$) quantifies the complementary knowledge that cannot be fully captured by the model $y$ but can be better understood by the other models.
Intuitively, a high CHR value means that the model can take huge benefits from the complementarity of the other models, whereas CHR$(\cdot)=0$ means that the model already captures all the knowledge of the other models.

We provide two statistical analyses using CHR.
\textbf{(1)} We compare CHR values induced by learning objective with values induced by two other factors, which influence the information that a model learns, investigated in recent work:
initialization \cite{DML} and model size (i.e., capacity) \cite{BD}.
Given a model $y$, we generate a set of models (including $y$) with five different conditions for each factor\footnote{
For the initial condition, we use the default initialization schemes provided in PyTorch and control it by setting different random seeds.
For the model size, we control the user/item embedding dimension $\in \{32, 48, 64, 80, 96\}$, and the size of subsequent layers are changed accordingly.
It is worth noting that different model sizes necessarily entail different initial conditions.}, i.e., $\mathcal{M}_{I}$ with different initial conditions, $\mathcal{M}_{C}$ with different capacities, and $\mathcal{M}_{O}$ with different learning objectives.
Then, we compute CHR for each factor, i.e., CHR($y;\mathcal{M}_{I}$), CHR($y;\mathcal{M}_{C}$), and CHR($y;\mathcal{M}_{O}$), and report the results in Table \ref{tbl:CHR}. 
\textbf{(2)} Given $\mathcal{M}_{O}$, we compute \textit{user-level} CHR\footnote{Eq.\ref{eq:chr} is computed for each user, i.e., ${|\bigcup_{x \in \mathcal{M}_{O}}\mathcal{H}^u_x - \mathcal{H}^u_y|}\, / \,{|\bigcup_{x \in \mathcal{M}_{O}} \mathcal{H}^u_x|}$, where $\mathcal{H}^u_x$ is the test interactions of user $u$ correctly predicted by model $x$.}, then we present CDF of CHR for the users in Figure \ref{fig:prelim_analysis}b.

\begin{table}[t]
\caption{CHR induced by three factors.
All values are computed for the same model $y$ having the identical initial condition, embedding dimension 64, and optimized by~CF-A.}
\renewcommand{\arraystretch}{0.7}
\renewcommand{\tabcolsep}{0.35cm}
\begin{tabular}{c ccc}
\toprule
Factors & CiteULike   & Ciao  & Foursquare   \\
\midrule
Initial condition & 0.30  & 0.23 & 0.29 \\
Model size & 0.32 & 0.28 & 0.38 \\
Learning objective & 0.56 & 0.43 & 0.54 \\
\toprule
\end{tabular} 
\label{tbl:CHR}
\vspace{-0.5cm}
\end{table}

\vspace{0.1cm}
\noindent
\textbf{Observations.}
\textbf{(1)} Learning objective incurs significantly high CHR (up to 75 \% higher than that from model size on CiteULike).
This indicates the learning objective is indeed an important factor affecting the knowledge captured by a model.
The other factors can be also exploited in the proposed framework to enhance the complementarity.
In this work, however, we focus on the learning objective which has not been studied well in the previous literature.
\textbf{(2)} Considerable complementarity exists for the majority of users, e.g., on CiteULike, most users have CHR larger than 0.4, which indicates that only less than 60\% of the total correct prediction set (i.e., $\bigcup_{x \in \mathcal{M}_o} \mathcal{H}_x$) is reached with a single learning~objective.

In sum, we conclude as follows:
(1) The learning objective is a key factor that determines the knowledge captured by the model.
(2) Models trained with different objectives learn distinct aspects of user-item relationships, which in turn produces complementary recommendations.
(3) The complementarity can be exploited to provide more accurate recommendations to more users, compared to the case of considering each single-faceted knowledge.

\section{PROPOSED FRAMEWORK---\proposed}
\label{sec:method}
We present our framework, named as \proposed (\underline{Con}sensus learning for OC\underline{CF}),
which exploits the complementarity from heterogeneous objectives throughout the training process, generating a more generalizable model (Fig.\ref{fig:overview}b).
We first provide an overview of \proposed (Section 3.1). 
Then, we introduce a consensus learning strategy from heterogeneous objectives (Section 3.2).
Lastly, we present our solution to balance the heterogeneous objectives without additional hyperparameters (Section 3.3).
The training details and Algorithm of \proposed are provided in Appendix~\ref{app:training}.

\vspace{-2pt}

\subsection{Overview}
In training, \proposed constructs a multi-branch variant of a given target model by adding auxiliary heads, each of which is trained with heterogeneous learning objectives. 
Then, it generates \textit{consensus} of multiple views from different heads, and it guides the heads based on the consensus.
The superscript $t$ denotes $t$-th training step (or epoch).
The overall loss function is formulated as:
\begin{equation}
    \begin{aligned}
        \mathcal{L}^t = \sum_{x \in \mathcal{F}} \lambda_{x}^t \mathcal{L}^t_{x},
    \end{aligned}
    \label{eq:L}
\end{equation}
where $\mathcal{F}$ is the set of the heads, $\mathcal{L}_x^t$ is the loss function for head $x$ at training step $t$, and $\lambda_{x}^t$ is the trainable parameter for dynamically balancing the heads during training.
Each head $x$ is trained with two loss terms: the original CF loss ($\mathcal{L}^t_{CF\text{-}x}$) and the consensus learning loss which aligns its prediction to the consensus ($\mathcal{L}^t_{CL\text{-}x}$). 
\begin{equation}
    \begin{aligned}
        \mathcal{L}^t_x =  \mathcal{L}^t_{CF\text{-}x} + \alpha \mathcal{L}^t_{CL\text{-}x}
    \end{aligned}
    \label{eq:Lx}
\end{equation}
where $\alpha$ is the hyperparameter controlling the effect of the consensus learning.
After training, we convert the multi-branch architecture back to the original single-branch model by removing the auxiliary heads, thus there is no additional inference cost for the deployment.
Further, the consensus from multiple heads can be also used as a high capacity model in the scenario where there is less constraint on the inference cost.

\vspace{-2pt}

\subsection{Collaborative Evolution with Consensus}
The consensus learning from the heterogeneous objectives raises several challenges.
First, the distributions and semantics of their prediction scores are different, thus the predictions from the heads need to be represented as a unified form so that they can learn interchangeably from each other.
For example, $\hat{r}_{ui}=0$ means a weak relation with CF-A, but a very strong relation with CF-B.
These discrepancies make the naive score-based knowledge exchange \cite{DML, BD, ONE} inapplicable.
Second, to generate informative consensus beneficial for all heads, it is essential to identify \textit{reliable} predictions from each head and selectively reflect them into the consensus.
As shown in Section 2.3, the user-item relationships correctly predicted by each objective are significantly different.
Without considering the reliability, the consensus will contain incorrect predictions, providing inaccurate supervision~for~the~heads.

We propose an effective strategy to resolve the challenges.
Instead of the score (i.e., $\hat{r}_{ui}$) itself, we utilize \textit{ranking information} (e.g., $\hat{r}_{ui} > \hat{r}_{uj} > \hat{r}_{uk} $).
These relative preference priorities are the universal form of knowledge that can be derived from all heads, enabling the knowledge exchange among the heads with different output distributions.
Further, we devise a new way of generating the consensus, considering both the predicted rankings and their reliability~revealed~from~\textit{temporal consistency}~of~the~predictions.

\subsubsection{Generating consensus}
For each user, we generate the item ranking list by sorting the prediction scores from each head.
Let $\pi^t_x$ denote the ranking list from head $x$ at training epoch $t$ ($u$ is omitted for simplicity).
We assess the importance of each item in terms of top-$N$ recommendation, then consolidate the importance from each head to generate the consensus.
Concretely, we generate the ranking consensus $\pi^t$ from $\{\pi^t_x \mid x \in \mathcal{F}\}$ by reranking items according to their overall importance.
The importance is assessed by two factors: \textit{ranking position} and \textit{consistency}.
The ranking position indicates the user's potential preference on each item, naturally, high-ranked items need to be considered important for the top-$N$ recommendation.
The consistency indicates how consistent the predicted ranking is during training, and according to our analysis, this is an important indicator to identify reliable~predictions.

\vspace{0.05cm}
\noindent
\textbf{Ranking position.}
Let $\text{rank}^{t}(x,i)$ denote the rank of item $i$ in $\pi^t_x$ where a lower value means a higher ranking position, i.e., $\text{rank}^{t}(x,i)=0$ is the highest rank.
The importance of item $i$ by ranking position is defined as follows:
\begin{equation}
\begin{aligned}
R^t_{x,i} = f(\text{rank}^{t}(x,i)),
\end{aligned}
\end{equation}
where $f$ is the monotonically decreasing function.
In this work, we use $f(k) = e^{-k/T}$ to put more emphasis on top positions where $T$ is a hyperparameter controlling the emphasis.

\vspace{0.05cm}
\noindent
\textbf{Consistency.}
To identify reliable predictions (i.e., correct top-ranked predictions) from each ranking list $\pi^t_x$, we define consistency of predicted ranking position as follows:
\begin{equation}
\begin{aligned}
C^t_{x,i} &= f(\text{std}[\text{rank}^{t-W}(x,i), \ldots, \text{rank}^{t}(x,i)])
\end{aligned}
\label{eq:c}
\end{equation}
where $\text{std}[\cdot]$ is the standard deviation.
We compute the consistency of recent predictions from $t-W$ to $t$, where $W$ is the window size.
This metric gives higher importance to more consistent predictions.

\begin{figure}[t]
\begin{subfigure}[t]{1.0\linewidth}
    \includegraphics[width=\linewidth]{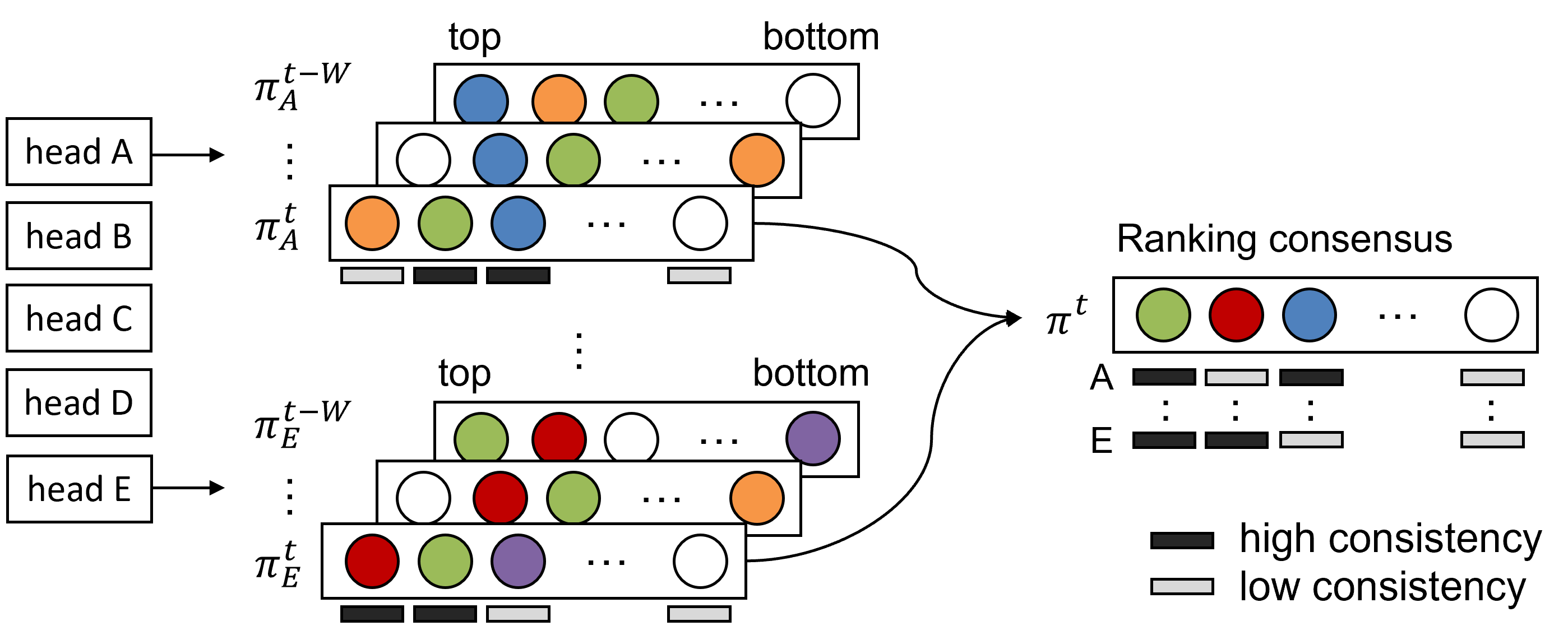}
\end{subfigure}
\caption{The conceptual illustration of the consensus generation process. Best viewed in color.}
\label{fig:consensus1}
\vspace{-0.5cm}
\end{figure}

\vspace{0.05cm}
Finally, the ranking consensus is generated by reranking items based on the consolidated importance.
For each item $i$, the importance $I^t_{i}$ is computed as follows:
\begin{equation}
\begin{aligned}
I^t_{i} = \mathbb{E}_{x \in \mathcal{F}}[RC^t_{x,i}], \,\,\, \text{where} \,\,\, RC^t_{x,i} = combine(R^t_{x,i}, C^t_{x,i}).
\end{aligned}
\end{equation}
$combine(\cdot,\cdot)$ is the function to consider both factors simultaneously. 
In this work, we use a simple addition.
The consensus generation process is illustrated in Figure \ref{fig:consensus1}.
The proposed strategy penalizes the frequently changed predictions, and pushes the items having not only high rank but also high consistency (green and blue in head A, red and green in head E) across the heads to the top of the ranking consensus $\mathcal{\pi}^t$.

\subsubsection{{Consensus learning}}
After generating the consensus, we enhance the performance of each head using the consensus.
The consensus can be thought of as the teacher model in the general two-stage knowledge distillation (KD) \cite{hinton2015distilling}.
However, unlike the general KD relying on the pretrained teacher that makes static predictions, the consensus collaborative evolves with the heads based on their complementarity, which can generate more accurate supervision beyond the static teacher.
That is, the improved heads generate more accurate consensus again, interactively boosting the recommendation quality throughout the training process.


We train each head $x$ to match its ranking orders with the consensus $\pi^t$.
To this end, we use the listwise learning-to-rank approach \cite{xia2008list-wise}. 
It defines a likelihood of the ranking order as a permutation probability based on the Plackett-Luce model \cite{marden2019analyzing} and trains the model to maximize the likelihood of the ground-truth ranking order.
We define the top-$N$ version of the permutation probability, which ignores the detailed ranking orders below the $N$-th rank, enabling each head to better focus on learning the top-$N$ ranking~results~\cite{DERRD}.
\begin{equation}
\begin{aligned}
p(\pi^t_{0:N}|\, \theta ) =  \prod_{{k}={1}}^{N} 
 \frac{{\exp} ( \hat{r}_{u, \pi^t(k)} ) }
 {\sum_{{i}= {k}}^{|\pi^t|}  {\exp} ( \hat{r}_{u, \pi^t(i)} )},
\end{aligned}
\label{eq:perm}
\end{equation}
where $\pi^t_{0:N}$ is the top-$N$ partial list, $\theta$ is the model parameters.
$\pi^t(k)$ is the $k$-th item of $\pi^t$, and $\hat{r}_{u, \pi^t(k)}$ is the relevance score for $(u, \pi^t(k))$ predicted by head $x$.
Then, we train each head to derive scores that agree with the consensus ranking by maximizing its log-likelihood.
The consensus learning loss is defined for users in the same batch $B$ used for the original CF loss as follows:
\begin{equation}
\begin{aligned}
 \mathcal{L}^t_{CL-x} = - \sum_{u \in B} \log p\left(\pi^{t, u}_{0:N} \mid \theta \right),
\end{aligned}
\label{eq:LCDx}
\end{equation}
where $\pi^{t, u}_{0:N}$ is the top-$N$ ranking consensus for user $u$.

\subsubsection{Empirical evidence}
\begin{figure}[t]
\hspace{-0.5cm}
\begin{subfigure}[t]{0.495\linewidth}
    \includegraphics[width=\linewidth]{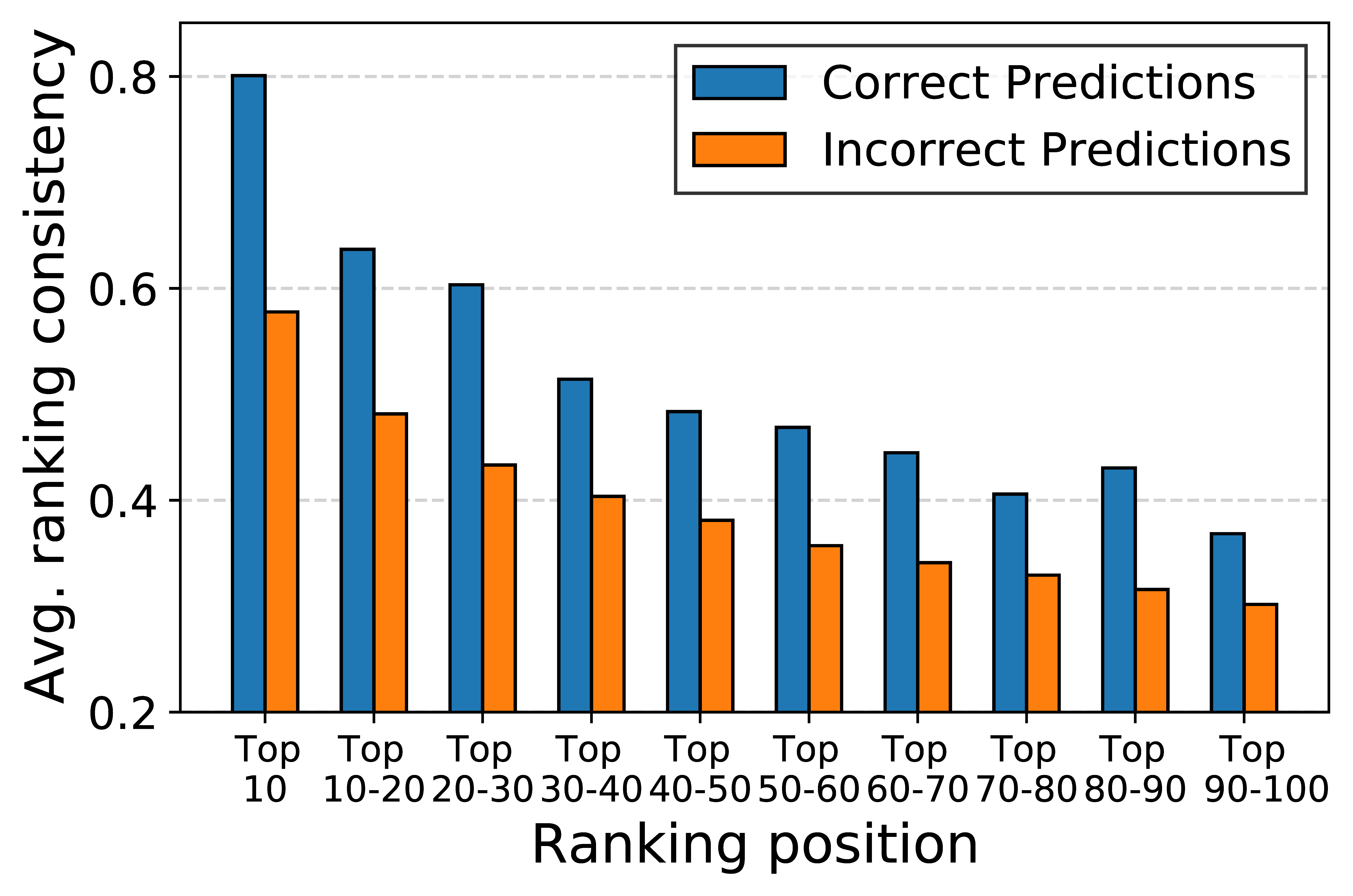}
\end{subfigure}
\hspace{-0.23cm}
\begin{subfigure}[t]{0.495\linewidth}
    \includegraphics[width=\linewidth]{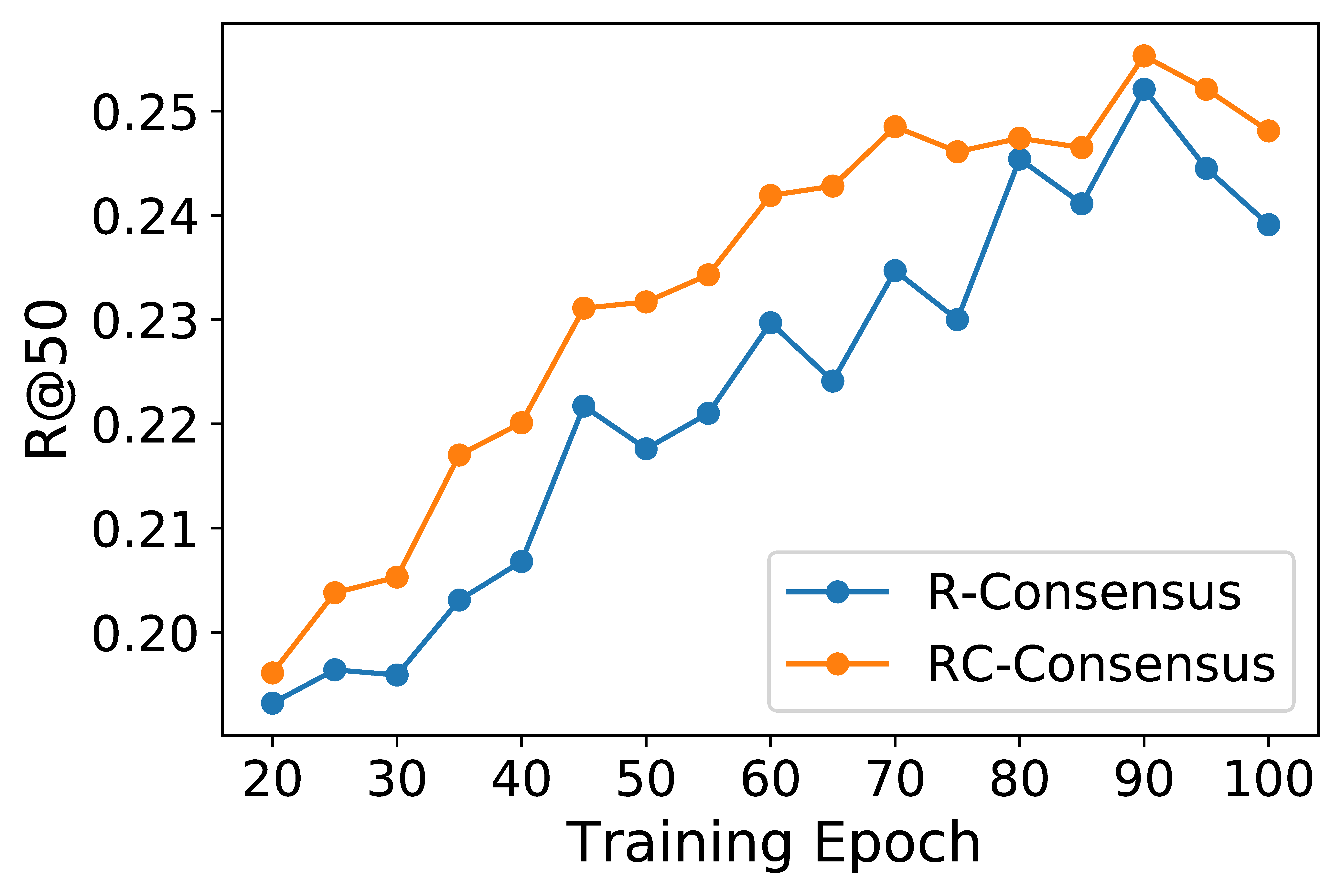}
\end{subfigure}
\caption*{\small (a) a single learning objective: $x \in \{A\}$}
\hspace{-0.55cm}
\vspace{-0.05cm}
\begin{subfigure}[t]{0.495\linewidth}
    \includegraphics[width=\linewidth]{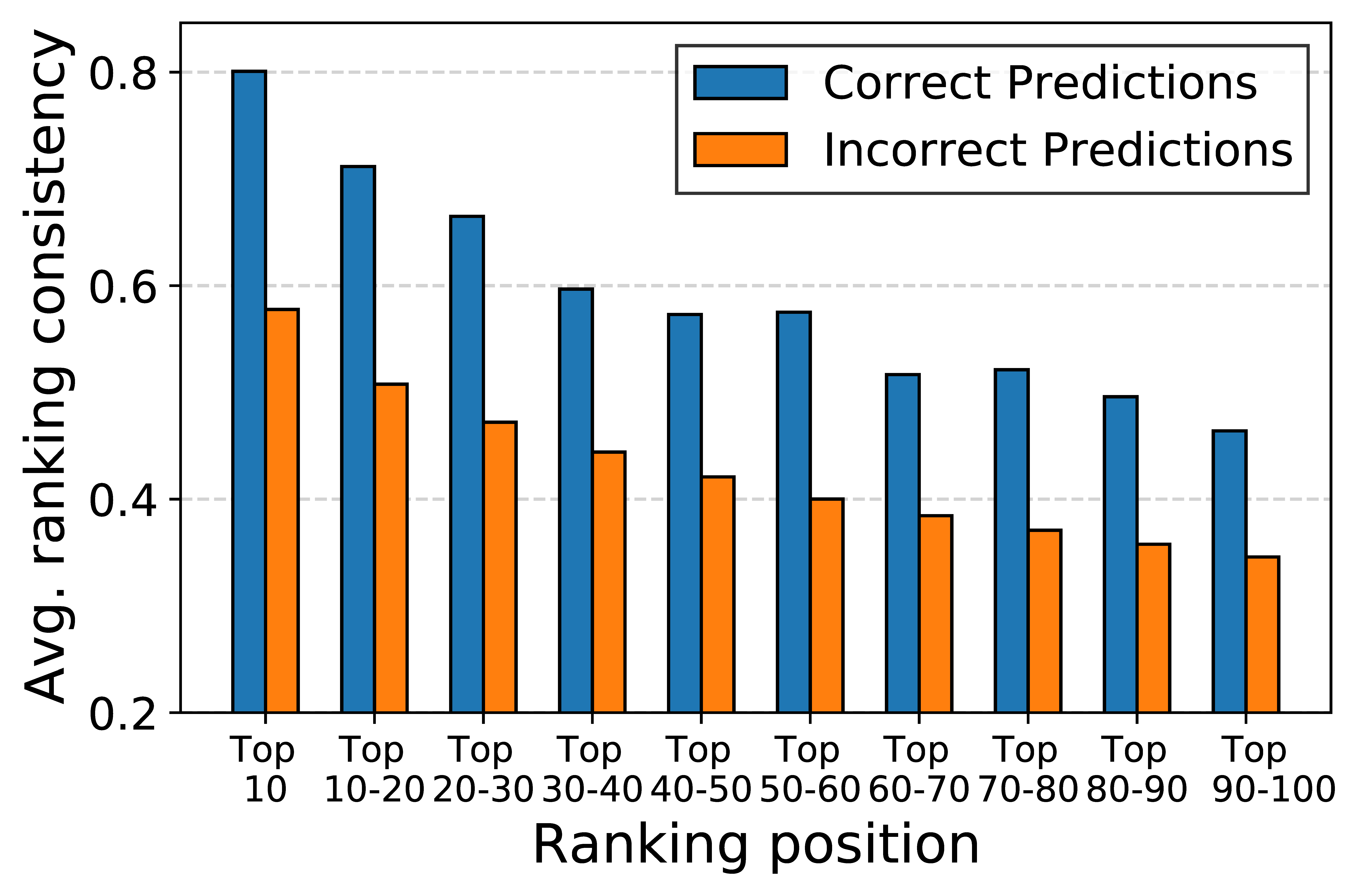}
\end{subfigure}
\hspace{-0.23cm}
\begin{subfigure}[t]{0.495\linewidth}
    \includegraphics[width=\linewidth]{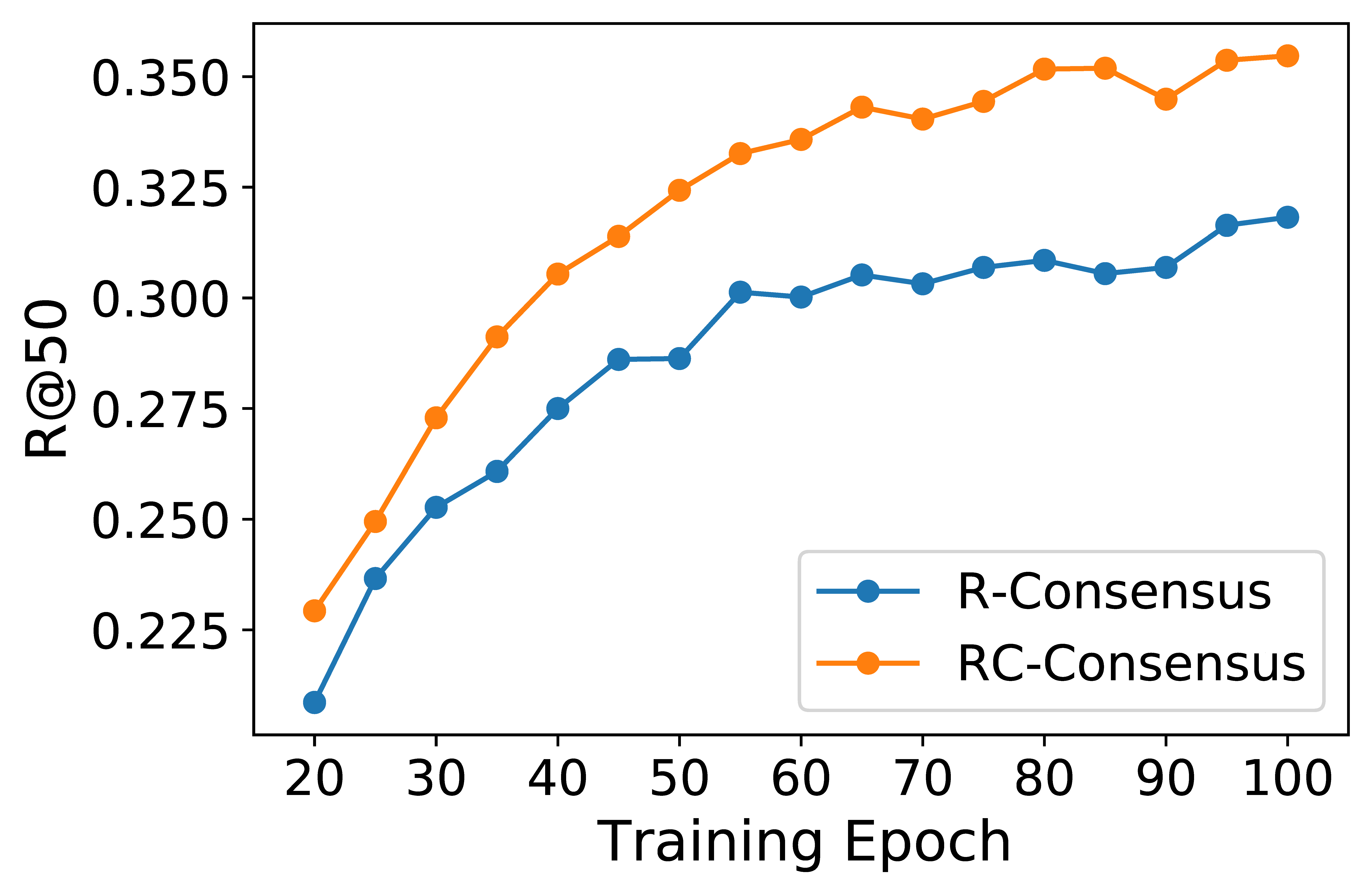}
\end{subfigure}
\hspace{-0.25cm}
\caption*{\small (b) multiple learning objectives: $x \in \{A, B, C, D, E\}$}
\caption{(left) Computed consistency, (right) Recall@50 of the generated consensus.}
\label{fig:consistency}
\vspace{-0.5cm}
\end{figure}

We investigate the reliability of the prediction measured by the consistency metric.
We analyze the dynamics of model predictions during training on CiteULike\footnote{The experimental setup is identical to Section 2.3.
The window size is set to 20.}.
First, we compare the computed consistency of correct predictions (i.e., test interactions) and that of incorrect predictions in the top-100 ranking list at epoch 50.
For each user's ranking list, we compute consistency for the item in each ranking position.
We then compute the average consistency of all users for each ranking position and summarize the results for 10 equispaced bins for better legibility. 
We report the results from CF-A (i.e., $C^{50}_{A,i}$) in Fig.\ref{fig:consistency}a (left) and the results from all objectives (i.e., $\mathbb{E}_{x}[C^{50}_{x,i}]$) in Fig.\ref{fig:consistency}b (left).

We observe that regardless of the learning objectives, for each rank bin
(1) the model tends to make consistent predictions on some user-item interactions but changes its prediction more frequently on others,
(2) the correct predictions have higher average consistency than the incorrect predictions. 
These observations indicate that the consistency reveals the reliability of each prediction and also show that even the items with the same rank need to have different importance based on the consistency in forming consensus.
Also, the model tends to make more consistent predictions for high-ranked items, indicating that the ranking consistency is a suitable indicator for assessing the reliability of the top-$N$ recommendation.

Further, we investigate whether consistency is indeed effective for generating accurate consensus.
In Figure \ref{fig:consistency} (right), we compare the accuracy of consensus generated by two importance criteria:
\textbf{(1) R-consensus} by $\mathbb{E}_{x}[R^t_{x,i}]$,
\textbf{(2) RC-consensus} by $\mathbb{E}_{x}[RC^t_{x,i}]$.
R-consensus can be thought of as a kind of majority voting scheme of ensemble;
items with the same rank are considered to have the same importance, and items with an overall high rank across the heads would be ranked near the top in the consensus.
As discussed earlier, this is insufficient to create an informative consensus.
It even generates the consensus more incorrect than the predictions of some models (reported in Section 4.1).
Note that in Fig.\ref{fig:consistency}a, R-consensus is equivalent to the original predictions by the model.

We observe that RC-consensus produces more accurate consensus than R-consensus (up to 16\% in Fig.\ref{fig:consistency}b).
Moreover, considering the consistency from all heads produces larger improvements compared to the case of a single head (Fig.\ref{fig:consistency}a vs. Fig.\ref{fig:consistency}b).
This again shows the efficacy of the proposed consensus generation strategy, which identifies and consolidates reliable predictions from~each~head.

\vspace{0.05cm}
\noindent
\textbf{Remarks.}
In sum, \proposed uses ranking information along with the consistency to create consensus, and guides the heads by using the ranking consensus.
Our strategy is inspired by the consistency regularization of semi-supervised learning (SSL) in image classification \cite{temporal_ensemble, tc-ssl} that deals with the uncertainty of unlabeled data by penalizing inconsistent predictions that change frequently during training.
Also, our observations are consistent with recent findings of SSL that frequently changed~predictions~are~unreliable~\cite{tc-ssl}.

\vspace{-2pt}
\subsection{Balancing Heterogeneous Objectives}
In \proposed, the heads are optimized by heterogeneous objectives having different convergence behaviors.
Thus, proper balancing of the heads is essential to effectively exploiting the complementarity.
If a few heads are too dominant during training, the shared parameters would be biased in favor of the heads, which incurs imbalances that impede the remaining heads from being fully optimized.
This degrades the overall effectiveness of the framework.

We enforce all the heads to be trained at similar rates by dynamically adjusting the gradient scale of each head on the shared parameters ($\theta_s$).
We apply gradient normalization technique \cite{gradnorm}, enabling the shared parameters to converge to beneficial features across all heads without additional hyperparameters for balancing the heads.
At each training step $t$, we define relative training ratio for each head $x$ as follows:
\begin{equation}
    \begin{aligned}
        \gamma^t_x = \frac{\mathcal{L}^t_x \, / \mathcal{L}^0_x}{\mathbb{E}_{x \in \mathcal{F}}[\mathcal{L}^t_x \,/ \mathcal{L}^0_x ]}, 
    \end{aligned}
\end{equation}
where $\mathcal{L}^0_x$ is the initial loss of the training. 
For $\gamma^t_x$, a lower value indicates head $x$ is optimized faster compared to the other heads. 
Based on the computed ratio, we adjust the gradient scale by each head.
For each head $x$, let $G^t_x = ||\nabla_{\theta_s} \lambda^t_x \mathcal{L}^t_x||_2$ denote the gradient scale by the loss of head $x$.
The target gradient scale is defined as $\mathbb{E}_{x\in \mathcal{F}}[G^t_x] \times \gamma^t_x$.
The gradient normalization is conducted by minimizing the distance between the gradient scales.
\begin{equation}
    \begin{aligned}
        \mathcal{L}^t_{b} = \sum_{x\in \mathcal{F}} \Big\vert G^t_x  - \mathbb{E}_{x\in \mathcal{F}}[G^t_x] \times \gamma^t_x \Big\vert.
    \end{aligned}
    \label{eq:LB}
\end{equation}
The target scale is treated as a constant, and $\mathcal{L}_{b}$ is differentiated only with respect to $\lambda^t_x$.
The computed gradient $\nabla_{\lambda^t_x} \mathcal{L}_{b}$ is then applied via standard gradient descent update rules to update $\lambda^{t+1}_x$.
Finally, $\lambda^{t+1}_x$ is normalized so that $\sum_{x\in \mathcal{F}} \lambda^{t+1}_x = 1$.

\section{Experiments}
\label{sec:experimentsetup}
\begin{table*}[t]
\small
\renewcommand{\arraystretch}{0.7}
\caption{Recommendation performances. \textit{Gain.Best} denotes the improvement of the best objective, which generates the best SingleCF model, in \proposed.
\textit{Gain.Con.} denotes the improvement of consensus over the best SingleCF model. We conduct the paired t-test with 0.05 level and all \textit{Gain}s are statistically significant.}
\renewcommand{\tabcolsep}{1.85mm}
\begin{tabular}{cc cccc cccc cccc}
\toprule
\multicolumn{2}{c}{Dataset}                      & \multicolumn{4}{c}{CiteULike}             & \multicolumn{4}{c}{Ciao}                  & \multicolumn{4}{c}{Foursquare}            \\
\cmidrule{1-2} \cmidrule(lr){3-6} \cmidrule(lr){7-10} \cmidrule(lr){11-14}
Training scheme & Objective  & R@20 & N@20 & R@50 & N@50 & R@20 & N@20 & R@50 & N@50 & R@20 & N@20 & R@50 & N@50 \\
\midrule
\multirow{5}{*}{\makecell{SingleCF}} 
& CF-A   & 0.1411    & 0.0870  & 0.2513    & 0.1160  & 0.1151    & 0.0766  & 0.1797    & 0.0952  & 0.1207    & 0.0897  & 0.1987    & 0.1151  \\
& CF-B  & 0.1551    & 0.0892  & 0.2851    & 0.1231  & 0.0645    & 0.0348  & 0.1547    & 0.0615  & 0.0897    & 0.0573  & 0.1746    & 0.0847  \\
& CF-C   & 0.1483    & 0.0774  & 0.2652    & 0.1086  & 0.1182    & 0.0784  & 0.1814    & 0.0964  & 0.1123    & 0.0786  & 0.1939    & 0.1052  \\
& CF-D    & 0.1739    & 0.1011  & 0.3132    & 0.1375  & 0.1042    & 0.0676  & 0.1755    & 0.0879  & 0.1026    & 0.0655  & 0.1948    & 0.0956  \\
& CF-E    & 0.1938    & 0.1125  & 0.3179    & 0.1451  & 0.0975    & 0.0539  & 0.1853    & 0.0797  & 0.1548    & 0.1123  & 0.2542    & 0.1457  \\
\midrule
\multirow{8}{*}{\proposed} 
& CF-A  & 0.2371          & 0.1378          & 0.3701          & 0.1734          & 0.1347          & 0.0845          & 0.2237          & 0.1134          & 0.1605          & 0.1115          & 0.2638          & 0.1451          \\
& CF-B   & 0.2350          & 0.1339          & 0.3733          & 0.1698          & 0.1350          & 0.0878          & 0.2213          & 0.1123          & 0.1720          & 0.1219          & 0.2773          & 0.1562          \\
& CF-C   & 0.2412          & 0.1395          & 0.3763          & 0.1756          & 0.1334          & 0.0880          & 0.2221          & 0.1131          & 0.1639          & 0.1125          & 0.2782          & 0.1495          \\
& CF-D   & 0.2300          & 0.1259          & 0.3716          & 0.1634          & 0.1349          & 0.0884          & 0.2266          & 0.1140          & \textbf{0.1757}          & 0.1235          & 0.2876          & 0.1579          \\
& CF-E  & 0.2418          & 0.1407          & 0.3721          & 0.1750          & 0.1367          & 0.0881          & 0.2248          & 0.1137          & 0.1714          & \textbf{0.1243}          & 0.2759          & \textbf{0.1584}          \\
& Consensus & \textbf{0.2533} & \textbf{0.1474} & \textbf{0.3896} & \textbf{0.1836} & \textbf{0.1420} & \textbf{0.0897} & \textbf{0.2303} & \textbf{0.1144} & 0.1749 & 0.1218 & \textbf{0.2877} & 0.1583 \\
\cmidrule{2-14}
& \textit{Gain.Best}  &	24.77\%	& 25.07\%	& 17.05\%	& 20.61\%	& 12.86\%	& 12.24\%	& 21.32\%	& 17.32\%	& 10.72\%	& 10.69\%	& 8.54\%& 8.72\% \\
& \textit{Gain.Con.}  & 30.70\%         & 31.02\%         & 22.55\%         & 26.53\%         & 20.14\%         & 14.41\%         & 24.28\%         & 18.67\%         & 12.98\%         & 8.46\%          & 13.18\%         & 8.65\%           \\
                                  \bottomrule
\end{tabular}
\label{tbl:main}
\vspace{-0.3cm}
\end{table*}

\begin{figure*}[t]
\centering
\hspace{-0.7cm}
    \begin{minipage}[r]{0.22\linewidth}
    \vspace{0.1cm}
        \centering
        \begin{subfigure}{1.17\textwidth}
          \includegraphics[width=1\textwidth]{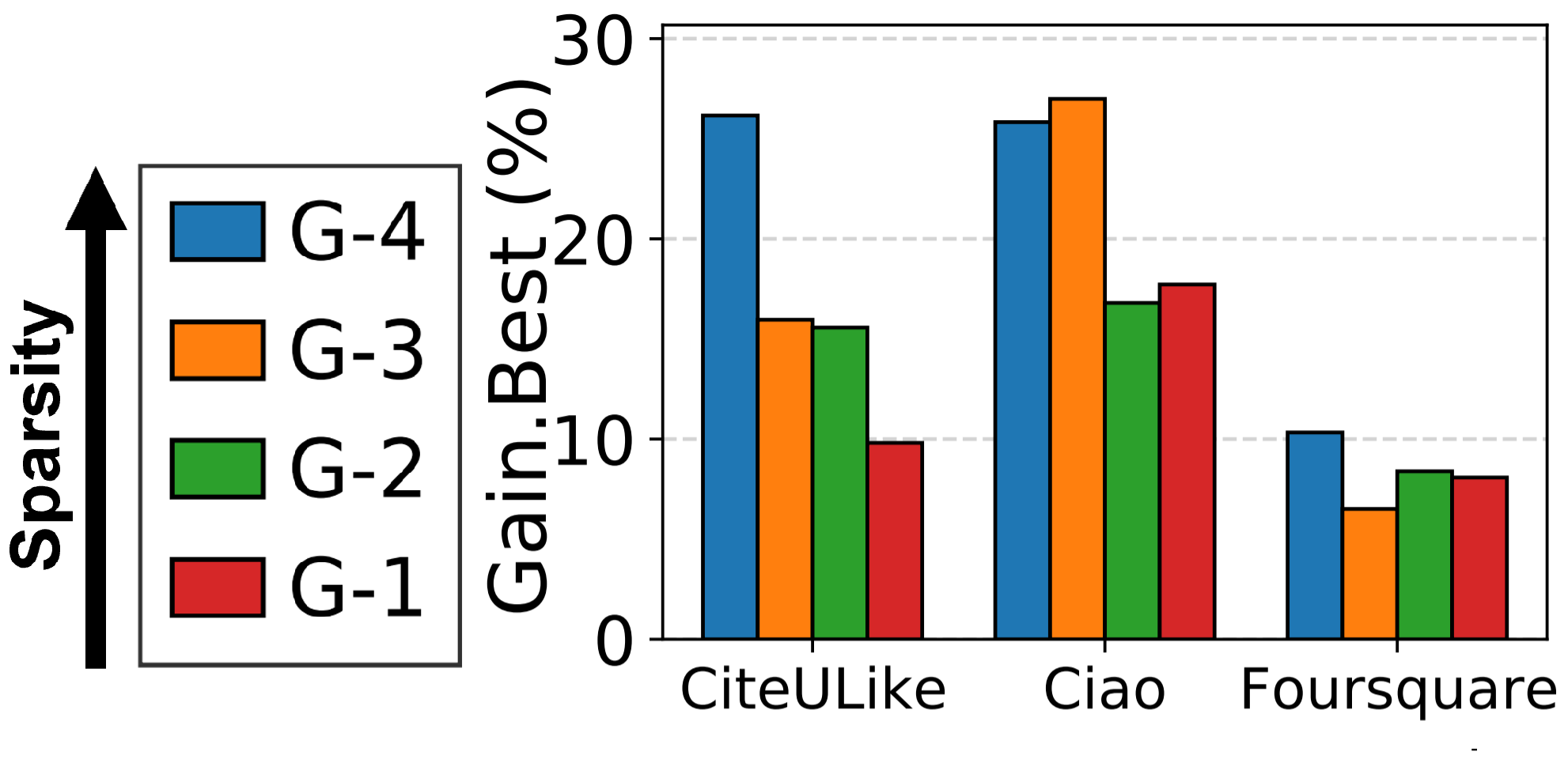}
        \end{subfigure}
        \vspace{0.01cm}
        \subcaption{\textit{Gain.Best} for user groups.}
    \end{minipage}
    \hspace{0.5cm}
    \begin{minipage}[r]{0.75\linewidth}
        \centering
        \begin{subfigure}{0.16\textwidth}
            \vspace{0.75cm}
          \includegraphics[width=1\textwidth]{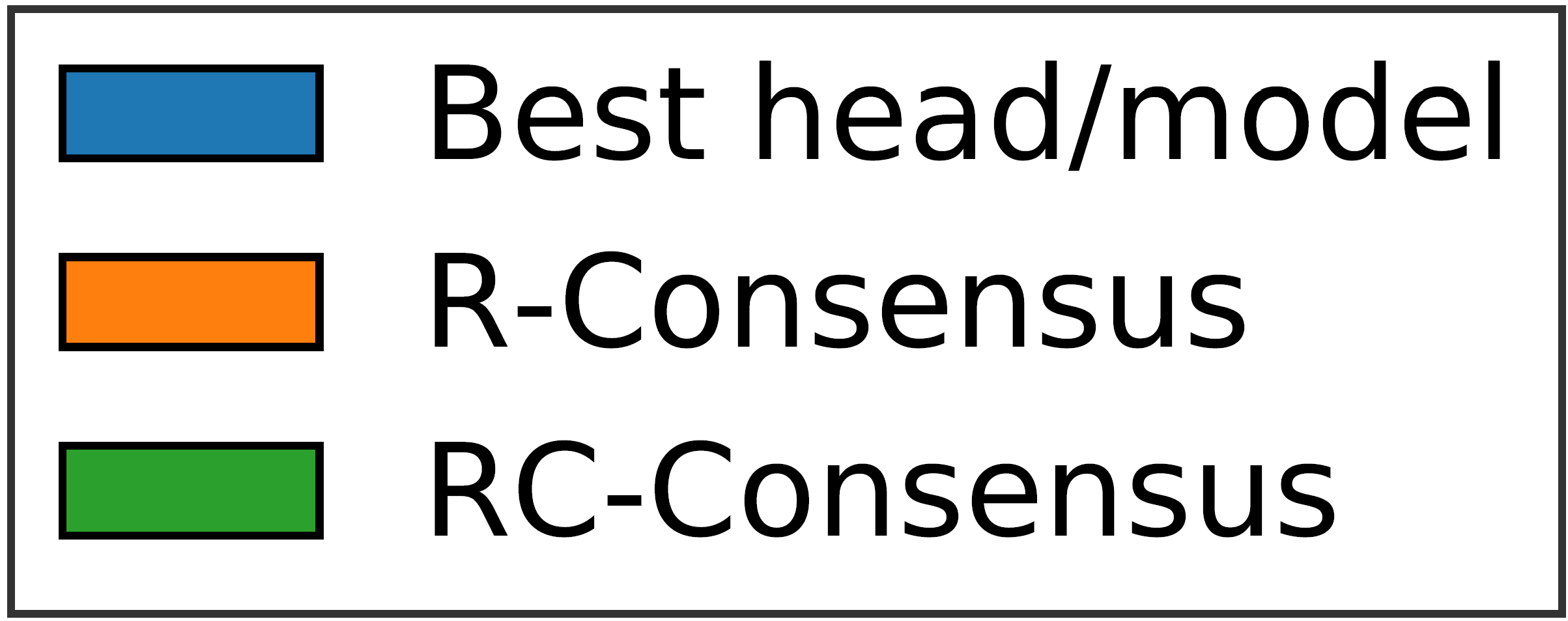}
        \end{subfigure}
        \hspace{-0.1cm}
        \begin{subfigure}{0.27\textwidth}
          \includegraphics[width=1\textwidth]{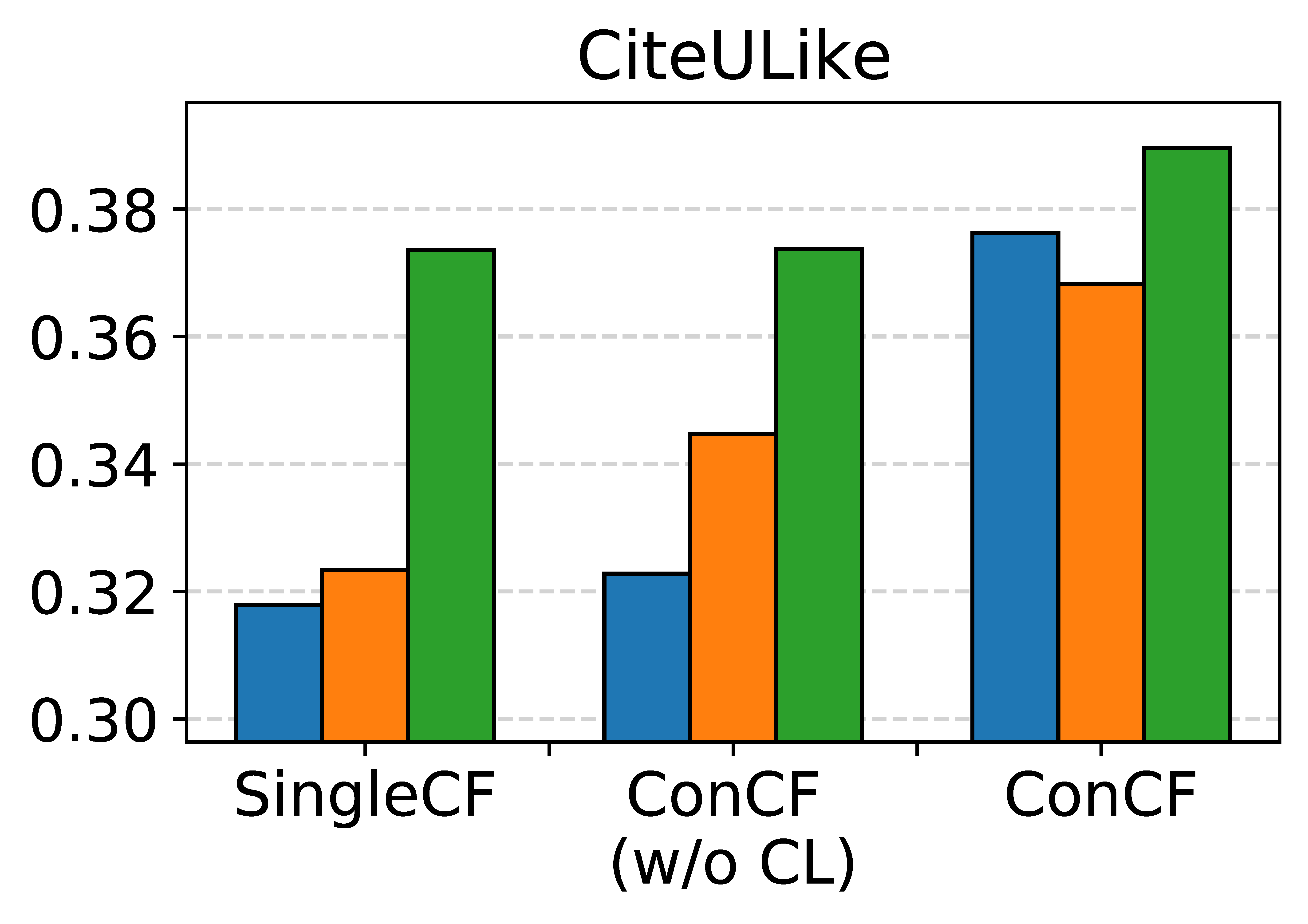}
        \end{subfigure}
        \hspace{-0.12cm}
        \begin{subfigure}{0.27\textwidth}
         \includegraphics[width=1\textwidth]{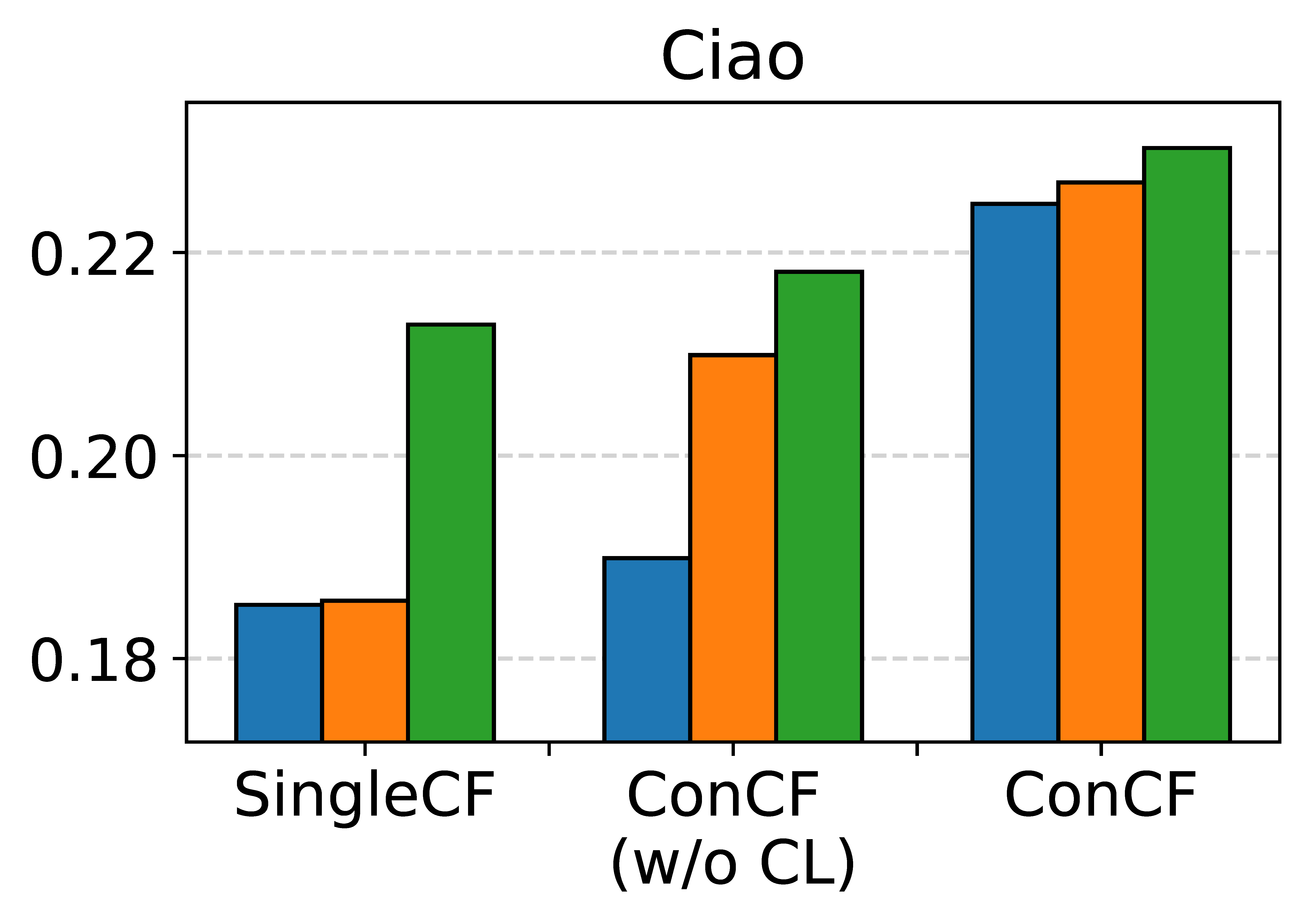}
        \end{subfigure}
        \hspace{-0.12cm}
        \begin{subfigure}{0.27\textwidth}
          \includegraphics[width=1\textwidth]{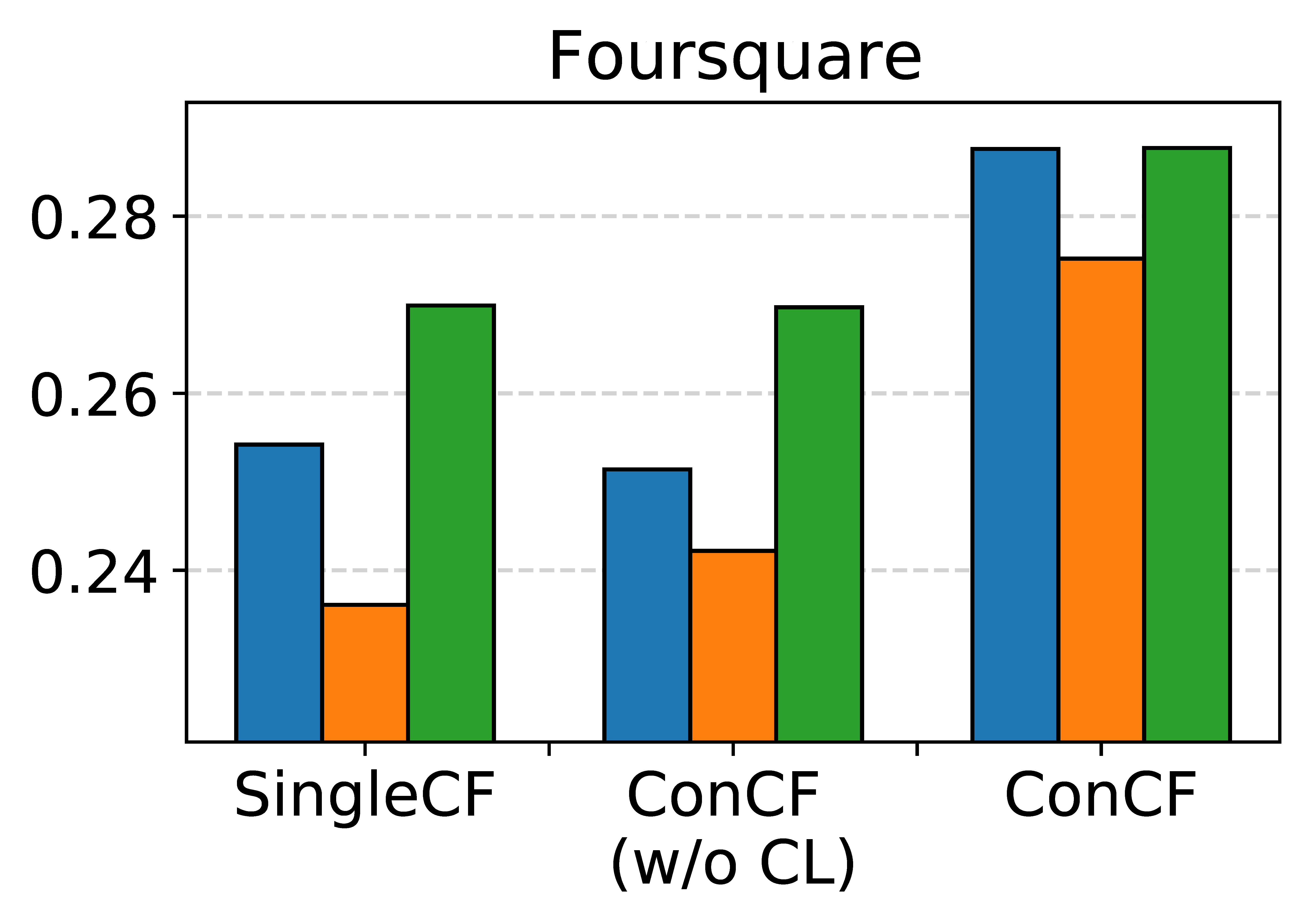}
        \end{subfigure}
        \vspace{-0.2cm}
        \subcaption{Comparison of three schemes. `ConCF (w/o CL)' is ConCF without consensus learning~loss.}
    \end{minipage}
    \hspace{-0.5cm}
    \caption{Further analyses on the consensus learning. All metrics are R@50.}
    \label{fig:further}
    \vspace{-0.3cm}
\end{figure*}

We provide a summary of the experiment setup due to limited space.
Refer to Appendix \ref{app:setup} for more detail.

\noindent
\textbf{Dataset.}
We use three real-world datasets: CiteULike \cite{wang2013collaborative}, Ciao \cite{tang2012mtrust}, and Foursquare \cite{liu2017experimental}.
These datasets are publicly available and widely used in recent work \cite{BUIR, DERRD, CML, BD}.

\noindent
\textbf{Evaluation protocol.}
We randomly split each user’s interaction history into train/valid/test sets in a 60\%/20\%/20\% split \cite{CML}.
We evaluate all models by two top-$N$ ranking metrics \cite{VAE, CML}: Recall@$N$ (R@$N$) and NDCG@$N$ (N@$N$). 
We report the average value of five independent runs, each of which uses differently split data.

\noindent
\textbf{Compared training schemes.}
\textbf{(a)} \bl uses a single objective for optimization as done in the most previous studies (Fig.\ref{fig:overview}a).
We train five independent models with each CF objective described in Section 2.2.
\textbf{(b)} \proposed uses \textit{multi-branch architecture} trained by \textit{heterogeneous objectives} along with~\textit{consensus learning} (Fig.\ref{fig:overview}b).
After training, we can make recommendations by using either 
1) a preferred head (e.g., based on the system requirement for further processing)
or 2) the consensus generated by all heads.
The former adds no extra inference cost compared to \bl and the latter can be adopted in the scenarios where computational cost is less constrained.
We report the results of both~cases~in~Table~\ref{tbl:main}.




\vspace{-0.1cm}
\subsection{Experiment Results}
\label{sec:result}
We provide various experiment results supporting the superiority of \proposed.
Supplementary experiments including hyperparameter study can be found in Appendix.

\subsubsection{Effectiveness of \proposed}
In Table \ref{tbl:main}, \proposed achieves significantly higher performance than all \bl models for all the datasets.
In particular, the performances achieved by each learning objective are considerably improved in \proposed beyond the best \bl~model.

We further analyze the impacts of \proposed on users with different sparsity levels.
We sort users in decreasing order according to their number of interactions (i.e., sparsity) and equally divide them into four groups; G-1 is the group of top-25\% users having many interactions, whereas G-4 is the group of users having few interactions.
We compute user-level $Gain.Best$ and report the average value for each user group in Fig.\ref{fig:further}a.
We observe that \proposed provides more accurate recommendations to all user groups, but particularly, the users with high sparsity (e.g., G-4) take more benefits from \proposed.
In \proposed, various aspects captured by the heterogeneous objectives provide a more complete understanding of the users, which can be particularly effective for the users with~very~limited~information.


\subsubsection{Consensus learning analysis.}
To validate the effectiveness of the consensus learning, we compare three training schemes: \bl, \proposed (w/o CL), and \proposed.
Since all these schemes can generate the consensus by consolidating the predictions from all models/heads, we report the performance of their best model (for \bl) or best head (for \proposed) as well as that of their consensus in Fig.\ref{fig:further}b.
Table \ref{tbl:chrvalue} presents CHR induced by each scheme.
In sum, \proposed achieves the best performance for both the best head and the final consensus (i.e., RC-con).
We analyze the results with various perspectives:

\begin{itemize}[leftmargin=*]
    
\item The consensus of the \bl models is not as effective as that of \proposed.
It can be seen as the \textit{ensemble model}; the predictions from multiple models are consolidated to make the final prediction.
Because the knowledge captured by a single objective is limited, each \bl model has limited performance.
This leads to unsatisfactory ensemble results despite the effectiveness of RC-con.

\item \proposed (w/o CL) cannot effectively improve the performance of each head, and also its consensus shows limited performance.
\proposed (w/o CL), which ablates the consensus learning (i.e., $\mathcal{L}_{CL}$) from \proposed, can be seen as conventional \textit{multi-task learning}.
Here, the knowledge of the objectives is exchanged only through the shared parameters without guidance on the predictions.
Its limited performance shows the importance of providing accurate and direct guidance on the predictions. 
Unlike the other two schemes, in \proposed, the heads learn reliable predictions from each other by negotiating their differences and reaching consensus throughout the training. 

\item RC-con is highly effective in generating consensus.
R-con sometimes even generates more inaccurate consensus than the best head/model (e.g., Foursquare), whereas RC-con always generates informative consensus beneficial for all heads.
This shows our strategy considering the reliability effectively consolidates the knowledge from heterogeneous objectives.

\item In Table \ref{tbl:chrvalue}, \proposed shows the lowest CHR values, which indicates the head effectively learns the complementary knowledge from the other objectives.
For \proposed (w/o CL) and ConCF, CHR is computed in terms of the head optimized by CF-A, i.e., CHR$(A;\mathcal{F})$, as done in the previous analysis (Table \ref{tbl:CHR}).
This result supports the superior performance of \proposed on Table \ref{tbl:main} and Fig.\ref{fig:further}b.
\end{itemize}

\begin{table}[t]
\caption{CHR by three training schemes.}
\small
\renewcommand{\arraystretch}{0.5}
\renewcommand{\tabcolsep}{3.5mm}
\begin{tabular}{cccc}
\toprule
Training schemes  & CiteULike & Ciao   & Foursquare \\
\midrule
\begin{tabular}[c]{@{}l@{}}SingleCF\end{tabular} & 0.56    & 0.43 & 0.54     \\
\begin{tabular}[c]{@{}l@{}}\proposed (w/o CL) \end{tabular} & 0.42    & 0.39 & 0.51     \\
\proposed & 0.18    & 0.10 & 0.17    \\
\bottomrule
\end{tabular}
\label{tbl:chrvalue}
\vspace{-0.5cm}
\end{table}

\subsubsection{Comparison with homogeneous objectives}
In Table \ref{fig:homo}, we analyze the importance of heterogeneous objectives in \proposed on CiteULike.
\proposed (w/o Het) is a variant of \proposed that uses the homogeneous objectives. 
We also present the results of PCL \cite{PCL} which is the state-of-the-art \textit{online KD} method for image classification.
It trains multiple classifier heads simultaneously by (multiclass) cross-entropy (i.e., CF-E) and guides the classifiers to minimize KL divergence from the ensemble Softmax logits that integrate the heads.
Here, the complementarity of the heads is mainly originated from the different initializations.

We observe that \proposed (w/o Het) still achieves considerably higher performance than the \bl model, however, it cannot achieve comparable performance to \proposed. 
These results are consistent with the findings of Table \ref{tbl:CHR} and again show the importance of exploiting the heterogeneous objectives.
Also, \proposed (w/o Het) performs better than PCL, which shows the effectiveness of utilizing ranking information for top-$N$ recommendation~\cite{RD, DERRD}.

\subsubsection{Comparison with two-stage KD}
In Table \ref{fig:kd}, we compare \proposed with conventional KD approach \cite{hinton2015distilling} on CiteULike.
Unlike \proposed where the consensus is dynamically updated during training, the teacher in two-stage KD makes static predictions.
We apply KD to \bl and \proposed (w/o CL).
Here, we use CF-E as it shows the best performance in the previous experiments.
\proposed (w/o CL) is guided by the static teacher via distillation.
All distillations are conducted in the same way using the listwise learning (Sec. 3.2.2).

We observe that the two-stage KD cannot generate the target model which significantly exceeds the teacher's performance.
Also, \proposed achieves the best performance among all training schemes.
In \proposed, the consensus collaboratively evolves with the heads based on their complementarity, which can generate more accurate supervision beyond the static teacher.


\begin{table}[t]
\small
\renewcommand{\arraystretch}{0.6}
\renewcommand{\tabcolsep}{2mm}
\caption{Comparison with homogeneous learning~objectives.}
\begin{tabular}{cccccc}
\toprule
Training scheme & Objective & R@20   & N@20   & R@50   & N@50   \\
\midrule
\multicolumn{1}{c}{\begin{tabular}[c]{@{}l@{}}SingleCF\end{tabular}} & CF$\text{-}$E          & 0.1938 & 0.1125 & 0.3179 & 0.1451 \\
\midrule
 & CF$\text{-}$E 1       & 0.2196 & 0.1350 & 0.3363 & 0.1662 \\
\multirow{4}{*}{PCL \cite{PCL}}  & CF$\text{-}$E 2        &0.2101 & 0.1291 & 0.3340 & 0.1621 \\
& CF$\text{-}$E 3        & 0.2133 & 0.1295 & 0.3471 & 0.1644 \\
& CF$\text{-}$E 4        &0.2139 & 0.1298 & 0.3344 & 0.1615 \\
& CF$\text{-}$E 5        & 0.2164 & 0.1318 & 0.3383 & 0.1635 \\
& Ensemble        & 0.2103 & 0.1331 & 0.3422 & 0.1674\\
\midrule
 & CF$\text{-}$E 1        & 0.2236 & 0.1369 & 0.3479 & 0.1689 \\
\multirow{4}{*}{\proposed (w/o Het)} & CF$\text{-}$E 2        & 0.2227 & 0.1400 & 0.3458 & 0.1718 \\
& CF$\text{-}$E 3        & 0.2282 & 0.1389 & 0.3495 & 0.1702 \\
& CF$\text{-}$E 4        & 0.2266 & 0.1400 & 0.3486 & 0.1712 \\
& CF$\text{-}$E 5        & 0.2256 & 0.1378 & 0.3441 & 0.1682 \\
& Consensus        & 0.2321 & 0.1407 & 0.3521 & 0.1731\\
\bottomrule
\end{tabular}
\label{fig:homo}
\vspace{-0.4cm}
\end{table}

\begin{table}[t]
\caption{R@50 comparison with two-stage KD. T1: the best \bl model (0.3179), T2: RC-con of all \bl models (0.3736).}
\small
\renewcommand{\arraystretch}{0.6}
\renewcommand{\tabcolsep}{0.75mm}
\begin{tabular}{ccccc}
\toprule
\multicolumn{1}{l}{}       & Teacher               &    Training scheme  & Best model/head & Consensus \\
\midrule
 & T1                                 & \bl (CF-E) + KD &    0.3207  &   -  \\
    two-stage  & T2               & \bl (CF-E) + KD &     0.3672  &-     \\
                           &      T2                              & ConCF (w/o CL)  + KD                   &    0.3704    &  0.3750   \\
\midrule
\multicolumn{1}{c}{one-stage} &  & ConCF                   &     0.3763  &  0.3896  \\
\bottomrule
\end{tabular}
\label{fig:kd}
\vspace{-0.4cm}
\end{table}

\vspace{-0.1cm}

\section{Related Work}
\label{sec:relatedwork}
We briefly introduce the work related to our study.
Recent OCCF methods are summarized in terms of their learning objectives in Section 2.2, and other related work including AutoML and multi-objective optimization are provided in Appendix \ref{app:related_work}.

\noindent
\textbf{Multi-task Learning.}
Multi-task learning (MTL) trains a single model that handles multiple tasks \cite{crawshaw2020multi, mmoe}.
MTL model learns commonalities and differences across different tasks, which can result in both improved efficiency and accuracy for each task \cite{mmoe}.
Extensive research has been conducted to improve the quality of MTL.
\cite{mmoe, tang2020progressive} have focused on designing the architecture for MTL, such as how to model task-specific and shared information.
\cite{gradnorm, liu2019loss} have focused on balancing the losses from multiple tasks.
\proposed also takes advantage of MTL, achieving better generalization and significantly reducing the parameters compared to training a separate model for each objective.
We also tried the various MTL architectures \cite{mmoe, tang2020progressive}, but no significant improvement is observed.

\vspace{0.05cm}
\noindent
\textbf{Knowledge Distillation.} 
Most KD methods \cite{hinton2015distilling, RD, DCD, DERRD, TD, IRRRD, PHR} have focused on model compression that transfers knowledge from a large pre-trained teacher model to improve a small target model.
Based on the extra supervision from the teacher, they have successfully improved the performance of the target model.
Despite their effectiveness on model compression, they require at least a two-stage training process and are also heavily dependent on the pre-trained teacher model \cite{DML, ONE}.
In computer vision, many studies \cite{DML, song2018collaborative, ONE, PCL, guo2020online} have focused on online KD which optimizes a target model by distilling knowledge among multiple models (or branches) in a single-stage training.
They consolidate the output from the multiple classifiers and guide the classifiers to follow the ensemble Softmax logits.
The ensemble logits integrate varying predictions of the classifiers that start from different initial conditions, which can help to improve the learning of the target model \cite{ONE}. 
For recommender system, \cite{BD} proposes an online KD method that simultaneously trains two models having different capacities, and \cite{zhang2020distilling} jointly trains an embedding-based model along with a meta-path based model for better explainability.

Although the aforementioned methods have improved performance by training multiple models simultaneously, they rely on a single objective to train the models.
Thus, they cannot exploit the complementarity from the heterogeneous objectives.
Also, they do not utilize ranking information for consolidating and guiding the models, which leads to sub-optimal ranking performance.

\vspace{-0.1cm}

\vspace{-0.1cm}

\section{Conclusion}
\label{sec:conclusion}
We first provide in-depth analyses showing the necessity of consolidating knowledge from heterogeneous OCCF objectives, which has not been studied well in the previous studies.
Motivated by the analyses, we propose a novel \proposed framework whereby the multiple heads trained by heterogeneous objectives collaboratively evolve based on their complementarity throughout the training process.
\proposed generates the ranking consensus by selectively reflecting the reliable predictions from each head, and enhances the heads by using the consensus.
The improved heads generate more accurate consensus again, interactively boosting the recommendation quality.
We validate the superiority of \proposed by extensive experiments.
Especially, the comparison with various training schemes shows the effectiveness of each proposed component.
We leave the study of exploiting the complementarity from multiple factors (e.g., model size, initialization) simultaneously for future work.

\vspace{0.1cm}
\noindent
\textbf{Acknowledgement.}
This work was supported by the IITP grant funded by the MSIT (No.2018-0-00584, No.2019-0-01906), the NRF grant funded by the MSIT (No.2020R1A2B5B03097210), the Technology Innovation Program funded by the MOTIE (No.20014926), 
and a grant of the Korea Health Technology R\&D Project through the KHIDI, funded by the MOHW (HI18C2383).

\pagebreak
\newpage
\clearpage
\bibliographystyle{ACM-Reference-Format}         
\bibliography{acmart}

\pagebreak
\newpage
\clearpage
\label{sec:appendix}
\appendix
\clearpage

\section{Appendix}

\subsection{Experiment Setup}
The source code of ConCF is publicly available through the author's GitHub repository\footnote{\url{https://github.com/SeongKu-Kang/ConCF_WWW22}}.

\vspace{0.2cm}
\label{app:setup}
\noindent
\textbf{Dataset.}
We use three real-world datasets: CiteULike \cite{wang2013collaborative}, Ciao \cite{tang2012mtrust}, and Foursquare \cite{liu2017experimental}.
These datasets are publicly available and widely used in recent studies \cite{BUIR, DERRD, CML}.
We follow the preprocessing of \cite{BUIR}.
Table~\ref{tbl:datastats} summarizes the statistics of the datasets.
\begin{table}[h]
\caption{The statistics of the datasets.}
\small
\renewcommand{\arraystretch}{0.9}
\renewcommand{\tabcolsep}{2.5mm}
\centering
\begin{tabular}{ccccc}
\toprule
Dataset & $\#$ Users  & $\#$ Items  & $\#$ Interactions & Density \\
\midrule
CiteULike  & 5,219  & 25,181  & 125,580       & 0.096\% \\
Ciao & 7,265  & 11,211  & 149,141        & 0.183$\%$ \\
Foursquare  & 19,465 & 28,593 & 1,115,108      & 0.200$\%$ \\
\bottomrule
\end{tabular}
\label{tbl:datastats}
\end{table}

\vspace{0.2cm}
\noindent
\textbf{Evaluation methodology.}
We randomly split each user’s interaction history into training/validation/test sets in a 60\%/20\%/20\% split \cite{CML}.
We set the maximum number of epochs to 500 and adopt the early stopping strategy;
it terminates when R@50 on the validation set does not increase for 20 successive epochs.
We keep the model with the best validation R@50 and report test set metrics with it.
Users and items that have less than 10 interactions are only included in the training set as done in \cite{CML}. 
As we focus on the top-$N$ recommendation task for implicit feedback, we evaluate all methods by using two widely used ranking metrics \cite{VAE, CML, SSCDR}: Recall@$N$ (R@$N$) and Normalized Discounted Cumulative Gain@$N$ (N@$N$). 
R@$N$ measures how many test items are included in the top-$N$ list and N@$N$ assigns higher scores on the upper-ranked test items.
We report the average value of five independent runs, each of which uses different random seeds for the data splits.

\vspace{0.2cm}
\noindent
\textbf{Model architecture.}
An overview of \bl and \proposed is illustrated in Figure \ref{fig:overview}.
In this work, we focus on the pure OCCF setting where the identity of each user (and item) is available \cite{NeuMF}.
The user-item encoder first maps the user/item id into embedding vector, and the embeddings are fed into subsequent layers \cite{NeuMF}.
Then, the head predicts a relevance score for each user-item pair by using the output representations of the user-item encoder.
We employ a two-layer perceptron with \textit{relu} activation for the user-item encoder and a single-layer perceptron for the head.
Note that a variety of modeling architectures can be flexibly adopted in the proposed framework.
We leave trying more diverse architectures including feature-based models and sequential models for future study.

\vspace{0.2cm}
\noindent
\textbf{Implementation details.}
We implement \proposed and all the baselines by using PyTorch and use the Adam optimizer to train all models.
The learning rate is set to 0.01 as it generally achieves the best performance for the separate networks.
The batch size is set to 1024.
For all models, we set the dimension size ($d$) of user/item embedding to 64 and [$d \rightarrow d/2 \rightarrow d/4$] for the encoder.
We uniformly sample a negative pair for each observed interaction as it shows good performance in recent work \cite{sun2020we, BUIR}.
We also notice that advanced sampling techniques \cite{NS_std} can be used, but it is not the focus of this work.
For \proposed, the importance of top positions in generating consensus ($T$) is set to 10, the importance of consensus learning loss ($\alpha$) is set to 0.01, and the target length of the recommendation list ($N$ in Eq.\ref{eq:perm}) is set to 50.
We apply CF-E on both row-wise and column-wise of the binary matrix $\mathbf{R}$.
The gradient normalization for balancing is applied to the shared user/item embeddings.
The details of the consistency computation are provided in A.2.


\begin{algorithm}[t]
\SetKwInOut{Input}{Input}
\SetKwInOut{Output}{Output}
\Input{Training data $\mathcal{D}$, \# total epochs $T_E$, Ranking prediction queue $Q$ with updating period $p$}
\Output{A trained target model $\{\theta_s, \theta_x\}$,\\ or a multi-branch model $\{\theta_s, \theta_x \text{ } \forall x\}$}
Initialize model parameters $\theta_{s}, \theta_{x}$ $\text{ } \forall x$\\
Initialize balancing parameters $\lambda^0_x = 1$ $\text{ } \forall x$\\
\BlankLine
\tcc{Training}
\For{$t=0,1,..., T_E$}{
\For{each batch $\mathcal{B} \in \mathcal{D}$}{
Compute $\mathcal{L}^t_{CF\text{-}x}$ $\text{ } \forall x$\\
\If{Warm-up ($t < \lvert Q \rvert \times p$)}{
    Compute $\mathcal{L}_x^t = \mathcal{L}^t_{CF\text{-}x}$ $\text{ } \forall x$\\
}
\Else{
    Compute $\mathcal{L}_x^t = \mathcal{L}^t_{CF\text{-}x} + \alpha \mathcal{L}^t_{CL\text{-}x}$ $\text{ } \forall x$ \COMMENT{\textit{Eq.\ref{eq:Lx}, \ref{eq:LCDx}}}\\
}
Compute $\mathcal{L}^t = \sum_{x \in \mathcal{F}} \lambda^t_x \mathcal{L}^t_x$ \COMMENT{\textit{Eq.\ref{eq:L}}}\\
Compute $\mathcal{L}^t_b$ and Update $\lambda_x$~$\text{ } \forall x$\COMMENT{\textit{Eq.\ref{eq:LB}}}\\
Update model parameters $\theta_{s}, \theta_{x}$ $\text{ } \forall x$
}
\If{$t \text{ } \% \text{ } p == 0$}{
    Compute ranking predictions and Update $Q$\\
    Generate consensus $\pi^{t}$ for all users
}
}
\BlankLine
\tcc{Testing}
Deploy with a target model $\{\theta_s, \theta_x\}$\\
Deploy with a multi-branch model $\{\theta_s, \theta_x \text{ } \forall x\}$
\caption{Algorithm of \proposed.}
\label{algo:algo}
\end{algorithm}

\subsection{Training Details}
\label{app:training}
The training procedure of \proposed is provided in Algorithm 1.
Also, we provide an analysis of hyperparameters for training in Table~\ref{tbl:hp}.

\vspace{0.2cm}
\noindent
\textbf{Efficient consistency computation.}
For generating the consensus, we compute the consistency of recent ranking predictions during training.
For an efficient implementation, we utilize a queue $Q$ to store the ranking predictions of $|Q|$ different epochs while gradually reflecting the latest predictions, then compute the consistency by using $Q$.
We update the queue and the consensus every $p$ epochs, so the window size $W$ can be thought of as $W=p\times|Q|$.
In this work, we set $p$ to 20, $|Q|$ to 5.
We provide the performance with different values in Table \ref{tbl:hp}.
We observe that \proposed achieves stable performance when the recent predictions are reflected more than a certain level (i.e., when the window size is large enough.).
Thus, we can efficiently compute the consistency by tracking the predictions of a few epochs ($|Q|$).
Interestingly, in the case of $|Q|=total\,\, epochs$, the performance is degraded, indicating that the recent prediction dynamics need to be considered important for the consistency computation.

\vspace{0.2cm}
\noindent
\textbf{Other details.}
For the model parameters, $\theta_s$ denotes the shared parameters for all heads (i.e., the user/item embedding, the user-item-encoder), and $\theta_x$ denotes the parameters of each head $x$ (line 1).
In the early stages of the training, we warm up the heads only with the original CF losses, ensuring enough prediction dynamics for generating consensus be collected and making each head sufficiently specialized for each CF objective (line 6-7).
For $\alpha$ which controls the effects of $\mathcal{L}^t_{CL}$, we observe the stable performance with $\alpha >= 0.01$ (Table \ref{tbl:hp}).
In this work, we set $\alpha=0.01$ (line 9).
Lastly, the balancing parameters $\lambda_*$ can be updated every batch or every epoch, we consistently obtain better results with the former.
As shown in Table \ref{tbl:hp}, the performance is degraded without balancing.

\subsection{Other Related Work}
\label{app:related_work}

\noindent
\textbf{AutoML and Loss function selection.}
In the field of AutoML, there have been a few attempts to automate the loss function search \cite{li2019lfs, SLF, AutoML}.
Recently, \cite{SLF, AutoML} select the most appropriate loss function among several candidate loss functions for each data instance by using Gumbel-Softmax.
There exists a fundamental difference between our work and the AutoML approach.
Our work is to \textit{consolidate} the complementary aspects from the heterogeneous objectives, whereas the AutoML approach \textit{selects} the most proper loss function \textit{in an exclusive manner.}
However, our analyses in Section 2.3 show that each objective captures different aspects of the user-item relationships, thus selecting a single objective provides an incomplete understanding.

To further ascertain our approach, we test the idea of exclusive selection.
\proposed-S is a variant of \proposed where each instance is trained by a single head (instead of all heads) selected by Gumbel-Softmax, and AutoML \cite{AutoML, SLF} trains each instance with a selected loss function without the consensus learning.
We observe that \proposed-S achieves better performance than \bl (Table \ref{tbl:main}), however, \proposed outperforms \proposed-S by a large margin.
For the AutoML approach, the performance of the best head is slightly degraded compared to \bl because it utilizes fewer training instances.
The ensemble result, which consolidates the predictions from all heads by the selection mechanism, shows a slightly better performance than \bl.
These results show that the exclusive selection is not effective enough to fully exploit the heterogeneous objectives, and again verify the effectiveness of our approach.

\vspace{0.2cm}
\noindent
\textbf{Multi-Objective Optimization.}
Multi-Objective Optimization (MOO) \cite{MOO, MOR2} aims to optimize more than one criterion (or desired goals) that may have trade-offs \cite{MOO}. 
Although the term `objective' serves as the main keyword in both MOO and our work, it is a distinct research direction from our work.
Unlike MOO which aims at generating a single model simultaneously satisfying multiple conflicting goals (e.g., accuracy and diversity), our work aims at consolidating the knowledge from the differently optimized models for the same criterion (i.e., ranking accuracy).
To this end, we utilize online consensus of separate heads (or models) having independent learning parameters, which raises novel challenges different from MOO.

\vfill\eject
\subsection{Supplementary Results}
\label{app:sup_result}
In Table \ref{tbl:sharing}, we compare different levels of parameter sharing among the objectives in \proposed on CiteULike.
`Full sharing' shares all parameters except for the heads, whereas `No sharing' shares no parameters, and each objective occupies a separate model.
Overall, parameter sharing between `full sharing' and `no sharing' shows good results, and in our experiment, sharing the user/item embeddings shows the best results. 
It is worth noting that the embeddings typically account for most parameters of OCCF model (more than 99\% on CiteULike). 

\begin{table}[t]
\caption{R@50 comparison with different~hyperparameters on CiteULike.}
\small
\renewcommand{\arraystretch}{0.9}
\renewcommand{\tabcolsep}{3.5mm}
\begin{tabular}{llcc}
\toprule
\multicolumn{2}{c}{Hyperparameters} & Best head & Consensus \\
\midrule[.1em]
\multicolumn{2}{l}{Without balancing (Section 3.3)}           & 0.3453    & 0.3575  \\
\midrule[.1em]
\multicolumn{2}{c}{$\alpha=0$}               & 0.3228    & 0.3747  \\
\multicolumn{2}{c}{$\,\,\,\,\,\,\,\alpha=0.01$}                  & 0.3763    & 0.3896  \\
\multicolumn{2}{c}{$\,\,\,\,\alpha=0.1$}                 & 0.3712    & 0.3836  \\
\multicolumn{2}{c}{$\alpha=1$}                & 0.3719    & 0.3823  \\
\midrule[.1em]
 & $p=1$                  & 0.3632    & 0.3661  \\
       $|Q|=total\,\, epochs$    & $p=10$                 & 0.3648    & 0.3685  \\
                         & $p=20$              & 0.3660     & 0.3664  \\
\cmidrule{1-4}
    & $p=1$                 & 0.3579    & 0.3691  \\
     $|Q|=5$        & $p=10$               & 0.3772    & 0.3872  \\
                         & $p=20 $                & 0.3763    & 0.3896  \\
\cmidrule{1-4}
   & $p=1$                  & 0.3708    & 0.3829  \\
     $|Q|=10$       & $p=10$                 & 0.3781    & 0.3907  \\
                         & $p=20$                 & 0.3779    & 0.3865 \\
\bottomrule[.1em]
\end{tabular}
\label{tbl:hp}
\end{table}

\begin{table}[t]
\small
\renewcommand{\arraystretch}{0.9}
\renewcommand{\tabcolsep}{2.5mm}
\caption{Performance comparison with loss function selection and AutoML approach on CiteULike.}
\begin{tabular}{ccccccc}
\toprule
                     &           & R@20    & N@20    & R@50    & N@50    \\
\midrule
\multirow{2}{*}{\proposed}          & Best head & 0.2412 & 0.1395 & 0.3763 & 0.1756 \\
                     & Consensus    & 0.2533 & 0.1474 & 0.3896 & 0.1836 \\
\midrule
\multirow{2}{*}{\proposed-S} & Best Head & 0.2158 & 0.1255 & 0.3487 & 0.1601 \\
                     & Consensus    & 0.2309 & 0.1351 & 0.3638 & 0.1750  \\
\midrule
\multirow{2}{*}{AutoML}  & Best Head & 0.1758 & 0.1032 & 0.3039 & 0.1371 \\
                     & Ensemble  & 0.1993 & 0.1204 & 0.3216 & 0.1529\\
\bottomrule
\end{tabular}
\end{table}

\begin{table}[h]
\caption{R@50 comparison of different parameter sharing.}
\small
\renewcommand{\arraystretch}{0.9}
\renewcommand{\tabcolsep}{2.6mm}
\begin{tabular}{ccccc}
\toprule
\multicolumn{1}{l}{}         &   Parameter sharing & Best head & R-con & RC-con \\
\midrule
\multirow{4}{*}{\proposed}    & Full sharing        & 0.3515            & 0.3512      & 0.3688       \\
                             & Embedding + 1 layer & 0.3702            & 0.3618      & 0.3805       \\
                             & Embedding           & 0.3763            & 0.3683      & 0.3896       \\
                             & No sharing      & 0.3662            & 0.3617      & 0.3836       \\
\bottomrule
\end{tabular}
\label{tbl:sharing}
\vspace{-0.55cm}
\end{table}

\end{document}